%% file: acl_latex.tex
\documentclass[11pt]{article}

\usepackage[preprint]{acl}

\usepackage{times}
\usepackage{latexsym}

\usepackage[T1]{fontenc}

\usepackage[utf8]{inputenc}

\usepackage{microtype}

\usepackage{inconsolata}

\usepackage{graphicx}

\usepackage{pifont}

\usepackage{booktabs}
\usepackage{multirow} 
\usepackage[table]{xcolor}
\usepackage{arydshln}

\usepackage{amsmath}
\usepackage{amssymb}
\usepackage{algorithm}
\usepackage{algpseudocode}

\usepackage{tcolorbox}
\tcbuselibrary{listings,skins}
\usepackage{capt-of}

\newcommand{\method}{ConRAG}

\title{\method{}: Consensus-Driven Evidence Flow for Multi-Hop\\ Question Answering}

\author{
  Yikai Zhu\thanks{Equal contribution.} \quad
  Kunfeng Chen\footnotemark[1] \quad
  Qihuang Zhong\thanks{Corresponding author.} \quad
  Juhua Liu \quad
  Bo Du \\
  School of Computer Science, Wuhan University \\
  Wuhan, China \\
  \texttt{\{yikaizhu,chenkunfeng,zhongqihuang,liujuhua,dubo\}@whu.edu.cn}
}

\begin{document}
\maketitle

\input{Sections/0_abstract}
\input{Sections/1_introduction}
\input{Sections/2_related_work}
\input{Sections/3_method}
\input{Sections/4_experiments}
\input{Sections/5_conclusion}
\input{Sections/6_limitations}
\input{Sections/7_ethics_statement}

\bibliography{reference}

\input{Sections/8_appendix}

\end{document}

%% file: Sections/0_abstract.tex
\begin{abstract}

Retrieval-augmented generation (RAG) has emerged as a promising paradigm for enhancing large language models (LLMs) on multi-hop question answering (QA), which requires reasoning over evidence from multiple documents.
However, current multi-hop RAG approaches still struggle with evidence selection and information transfer, \textit{i.e.}, retrieving irrelevant passages or failing to pass intermediate information to subsequent reasoning steps, thereby hindering coherent multi-hop reasoning.
To this end, we propose \textbf{\method{}}, a consensus-driven evidence flow framework that effectively boosts LLM performance on complex multi-hop QA.
In short, the core of \method{} is to select relevant evidence via consensus-driven multi-view ranking, and leverage a slot-based mechanism to transfer intermediate information between dependent reasoning steps.
Extensive experiments across three multi-hop QA benchmarks show that \method{} surpasses the baselines by a clear margin.
Encouragingly, on \texttt{Gemma-4-31B-it}, \method{} outperforms the strongest baseline by 5.8 points in average score, while reducing total end-to-end runtime and token consumption by 75.9\% and 67.3\%, respectively.

\end{abstract}

%% file: Sections/1_introduction.tex
\section{Introduction}

Retrieval-augmented generation (RAG) expands the knowledge boundaries of large language models (LLMs) by retrieving query-relevant evidence from external corpora~\citep{lewis2020rag,fan2024survey}.
Despite achieving remarkable performance on various knowledge-intensive tasks, RAG systems still fall short in complex multi-hop question answering (QA), which requires integrating complementary evidence distributed across documents.
Specifically, later hops may depend on bridging entities or relations revealed by intermediate evidence but absent from the original question~\citep{trivedi2023ircot}.
However, conventional RAG typically retrieves evidence once from the original query, limiting access to passages whose relevance emerges during reasoning.

\begin{figure}[t]
\centering  
\includegraphics[width=\columnwidth]{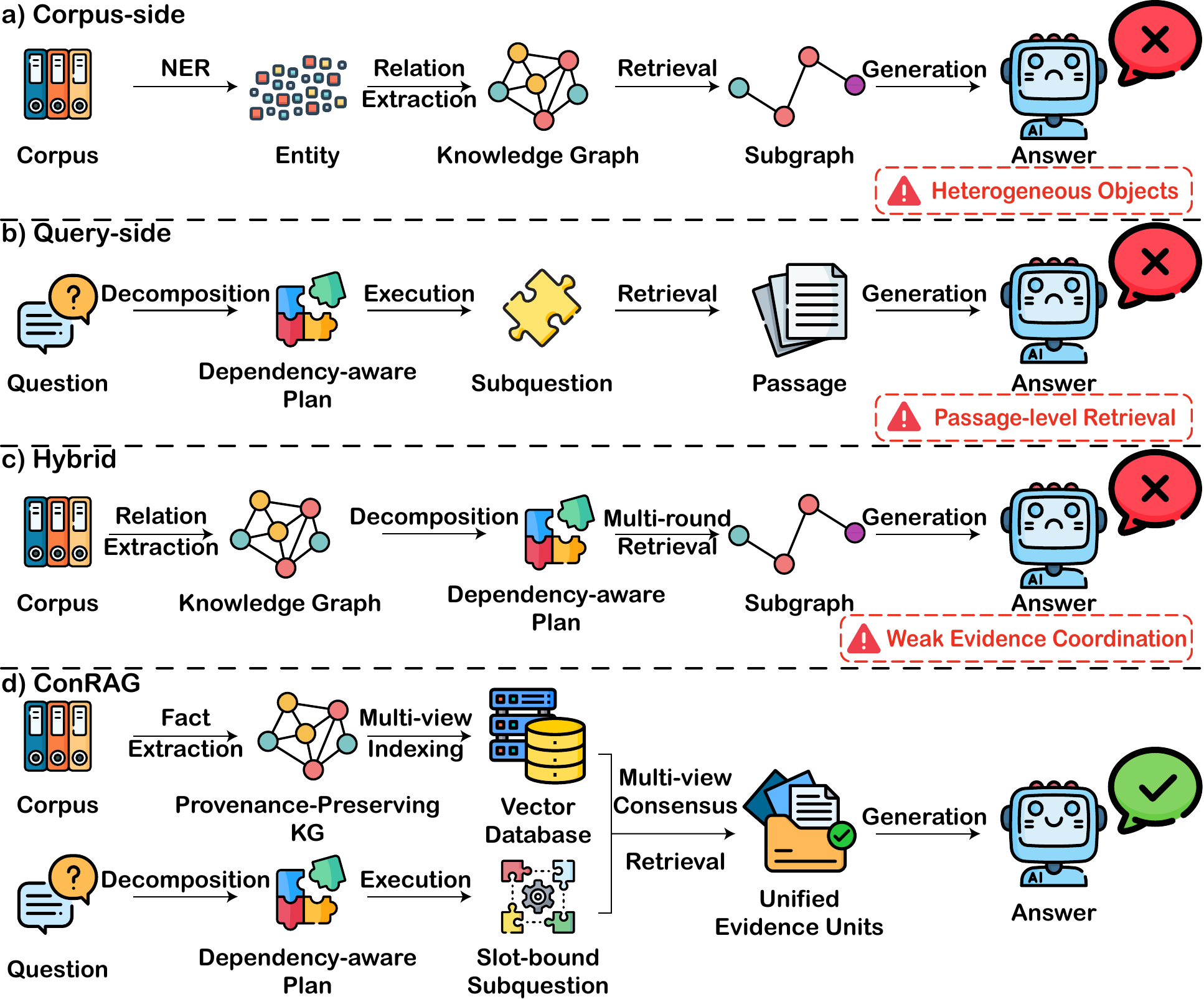}
  \caption{\textbf{Overview of different representative RAG paradigms} for multi-hop question answering.}
  \label{fig:overview_of_rag}
\end{figure}

Hence, multi-hop RAG has attracted considerable attention recently. As shown in Figure~\ref{fig:overview_of_rag}, existing research has primarily advanced along three main directions: 1) \textit{corpus-side} approaches, 2) \textit{query-side} approaches, and 3) \textit{hybrid} approaches.
Specifically, corpus-side approaches organize passages, entities, relations, or communities into graph-based or hierarchical retrieval spaces, enabling retrieval to exploit cross-passage connections and multi-granularity context~\citep{edge2024graphrag,guo2025lightrag,gutierrez2025hipporag2}.
Query-side approaches interleave retrieval with generated reasoning or learn to construct subsequent retrieval queries from intermediate evidence~\citep{trivedi2023ircot,zhuang2024efficientrag,chen2026logicrag}.
Hybrid approaches operate over structured corpus representations while conditioning subsequent evidence selection on intermediate retrieval or reasoning states~\citep{wang2024kgp,zhu2025chainrag,dong2026youtugraphrag}.
Despite their effectiveness, evidence selection and information transfer remain two central challenges in multi-hop RAG.
First, textual and structural retrieval signals can complement each other, but structural noise can interfere with evidence selection.
Second, ineffective transfer of intermediate information can hinder subsequent retrieval and disrupt multi-hop reasoning.
Thus, this raises a question: \textit{how can multi-hop RAG select complementary evidence and maintain coherent evidence flow across dependent reasoning steps?}

To this end, we propose \textbf{\method{}}, a consensus-driven evidence flow framework that coordinates evidence selection with intermediate information transfer through three interdependent operations: \textit{Connection}, \textit{Consensus}, and \textit{Constraint}.
On the corpus side, \textit{Connection} links recurring entities and relations across passages and preserves their source provenance to build a shared passage space for retrieval. During retrieval, \textit{Consensus} combines textual, entity, and relation evidence in this space and applies a bounded structural bonus. It also reduces the influence of entities occurring in many passages and diffuse relation matches. On the query side, \textit{Constraint} fills predefined query slots with answers from predecessor steps, runs steps with resolved dependencies concurrently, and limits information transfer to direct dependencies. Notably, \method{} uses the same \textit{Consensus} to retrieve evidence for both original queries and queries filled with answers from previous steps.

We evaluate \method{} on three popular multi-hop QA benchmarks, HotpotQA~\cite{yang2018hotpotqa}, 2WikiMultiHopQA~\cite{ho2020twowiki}, and MuSiQue~\cite{trivedi2022musique}.
Extensive experiments with \texttt{GPT-4o-mini} and \texttt{Gemma-4-31B-it} show that \method{} consistently improves over Vanilla RAG by a clear margin, \textit{i.e.}, up to \textbf{32.1} points in average score across all benchmarks. Encouragingly, on \texttt{Gemma-4-31B-it}, \method{} outperforms the strongest baseline by 5.8 points in average score and substantially reduces end-to-end computational costs. To summarize, our contributions are three-fold. (1) We propose \method{}, a consensus-driven evidence flow framework that effectively boosts the multi-hop RAG performance. (2) \method{} introduces two simple-yet-effective strategies, \textit{i.e.}, consensus-based retrieval and slot-based reasoning, to coordinate evidence selection and intermediate information transfer. (3) Extensive experiments prove that \method{} improves multi-hop QA performance and reduces total computational costs against the strongest baseline.

%% file: Sections/2_related_work.tex
\section{Related Work}

\paragraph{Corpus Structure and Adaptive Retrieval.}
Conventional RAG scores passages independently within a flat index~\citep{lewis2020rag,karpukhin2020dpr}. To improve multi-hop QA performance, subsequent research has focused on corpus organization and query adaptation. On the corpus side, methods construct hierarchical or graph-based evidence spaces through summaries, entities, and relations~\citep{sarthi2024raptor,edge2024graphrag,guo2025lightrag,wu2026memgraphrag}, and retrieve evidence through graph propagation or diffusion~\citep{gutierrez2024hipporag,gutierrez2025hipporag2,wang2026hgrag,zhuang2026linearrag}. On the query side, methods adapt retrieval through query decomposition, reasoning-guided interaction, and next-hop query construction~\citep{zhou2023leasttomost,yao2023react,trivedi2023ircot,zhuang2024efficientrag}. Some methods also use directed acyclic graphs (DAGs) to coordinate tool calls and reasoning steps~\citep{kim2024llmcompiler,chen2026logicrag}. In general, these approaches primarily improve corpus-side evidence organization or query-side retrieval adaptation. However, improvements on either side alone do not necessarily ensure consistent evidence selection and answer transfer across reasoning steps in multi-hop QA scenarios.

\paragraph{Cross-Stage Evidence Coordination.}
Hybrid systems bridge these two directions by combining structured corpus representations with adaptive retrieval. KGP maps model-generated evidence descriptions to neighboring passages in a document graph~\citep{wang2024kgp}. ChainRAG combines sentence-graph retrieval with subquestion rewriting based on previous answers~\citep{zhu2025chainrag}. Youtu-GraphRAG integrates schema-guided corpus construction with agentic query decomposition and multi-route retrieval~\citep{dong2026youtugraphrag}. However, selecting evidence from different sources to meet changing information needs across reasoning steps remains challenging. To address this challenge, \method{} combines consensus-driven multi-view ranking with explicit answer transfer through query slots to coordinate evidence flow across dependent reasoning steps.

%% file: Sections/3_method.tex
\section{Method}
\label{sec:method}

\begin{figure*}[t]
    \centering
    \includegraphics[width=\textwidth]{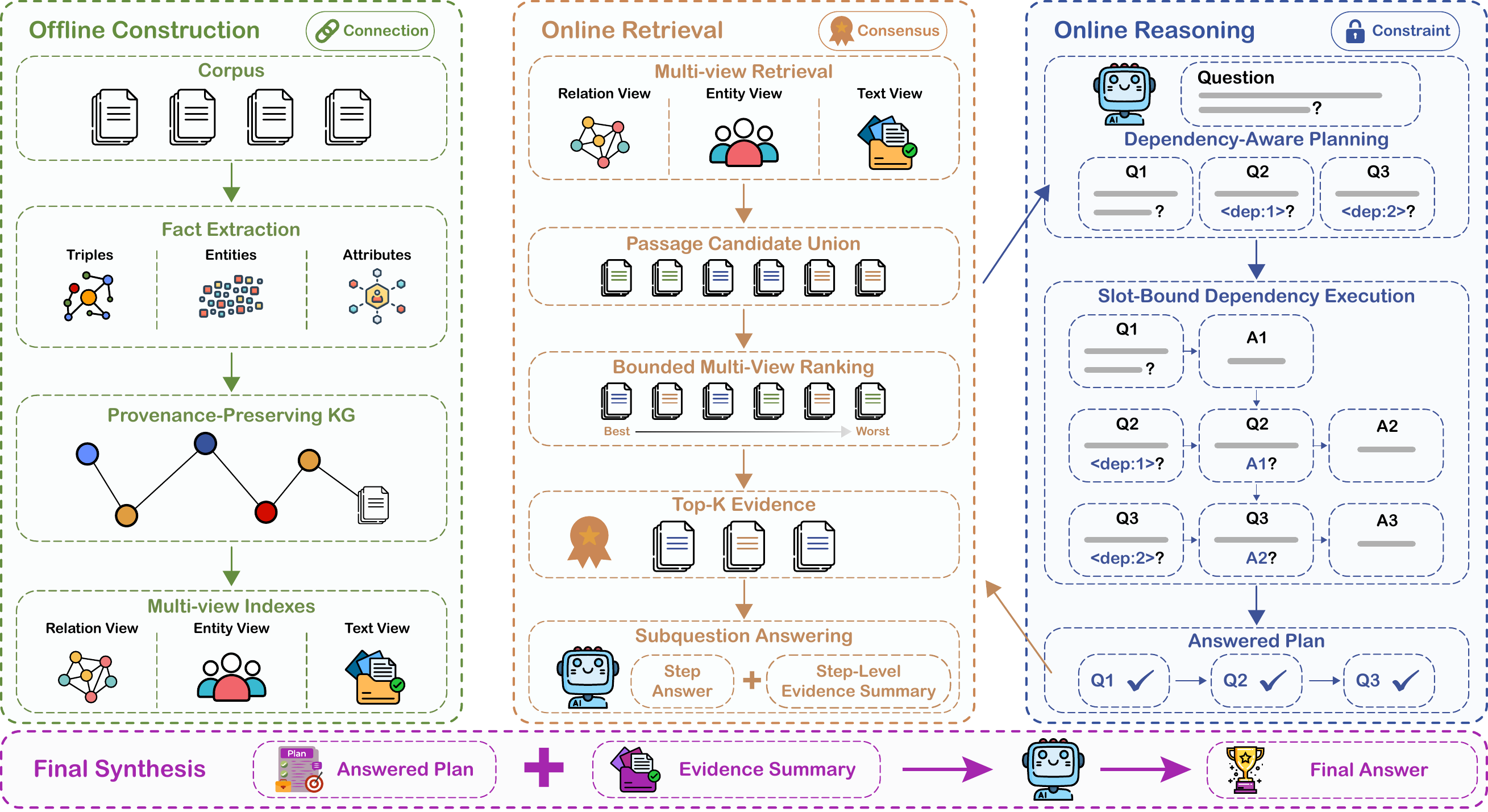}
    \caption{\textbf{Overview of \method{}}, which contains three main processes. \textit{Connection} links evidence across passages and preserves its sources. \textit{Consensus} combines Text, Entity, and Relation evidence for passage ranking. \textit{Constraint} transfers answers through query slots and executes independent steps concurrently.}
    \label{fig:framework}
\end{figure*}

Given a corpus of $n$ passages and a multi-hop question $q$, \method{} coordinates evidence selection and intermediate information transfer through three operations. As shown in Figure~\ref{fig:framework}, \textit{Connection} organizes evidence across passages, \textit{Consensus} ranks passages using three complementary views, and \textit{Constraint} guides retrieval and reasoning along query dependencies. The pseudo-code is provided in Algorithms~\ref{alg:connection}--\ref{alg:constraint} in Appendix~\ref{appendix_algorithms}.

\subsection{\textit{Connection}: Cross-Passage Organization}
\label{sec:connection}

Multi-hop QA requires connecting evidence scattered across passages. To support this process, \textit{Connection} merges shared entities and relations into a global directed multigraph and retains their source passages. For each passage $p_c$, a schema-guided extraction LLM identifies entities, attributes, and directed relations. Entities are merged by normalized names and types. Relations are merged by ordered keys containing the head entity, relation, tail entity, and both endpoint entity types. Each merged entity retains the unique values under each attribute key. The resulting graph contains entity nodes with attributes and labeled directed edges. Due to space limitations, more construction details are provided in Appendix~\ref{appendix_algorithms}.

We record the source passage IDs of each entity $e$ and relation $\rho$ as $\Pi_E(e)$ and $\Pi_R(\rho)$, respectively. These provenance sets allow each graph object to support retrieval of every passage in which it appears. We build three indices from $\ell_2$-normalized embeddings of passages, serialized entities with attributes, and serialized relation triples. Text hits identify passages directly. Entity and Relation hits map to passages through their provenance sets, placing all three views in a shared passage space for \textit{Consensus} ranking.

\subsection{\textit{Consensus}: Multi-View Retrieval}
\label{sec:consensus}

Structural evidence can complement Text relevance, but entities shared by many passages and numerous weak relation matches can receive excessive weight. To address these issues, \textit{Consensus} adjusts Entity and Relation support separately and adds a bounded structural bonus to the Text score. That is, \textit{Consensus} combines Text relevance with Entity and Relation support in a shared passage space.
Specifically, for query $q$, we retrieve up to $K_c$ objects from each index in parallel and map them to passages. Provenance projection can yield more than $K_c$ passages per structural view. Let $\operatorname{sim}(q,d)$ denote cosine similarity between the normalized embeddings of $q$ and document $d$. For each view, let $x_c$ denote the raw score of candidate passage $c$. We normalize these scores as $\operatorname{Norm}(x_c)=(x_c-x_{\min})/(x_{\max}-x_{\min})$, where $x_{\min}$ and $x_{\max}$ are the minimum and maximum scores among candidates in that view. If all scores are equal, every candidate receives a normalized score of one. An empty candidate set remains empty. We process the three views as follows.

\textbf{(1) Text normalization.} For Text hits with finite similarity scores, the base score on the Text view is normalized as $T_c=\operatorname{Norm}(\operatorname{sim}(q,p_c))$.

\textbf{(2) Entity weighting.} Entities shared by fewer passages provide more specific retrieval signals. We refer to this property as Entity specificity. Let $f_e=|\Pi_E(e)|$ be the number of passages containing entity $e$. Following inverse document frequency (IDF)~\citep{robertson2009bm25}, we define its specificity weight:
\begin{equation}
\sigma(e)=
\frac{\log\!\left(1+\dfrac{n-f_e+0.5}{f_e+0.5}\right)}
{\log\!\left(1+\dfrac{n-0.5}{1.5}\right)}.
\label{eq:entity_specificity}
\end{equation}
For each passage $c$, let $w_{ce}$ be the query similarity of a retrieved entity's serialized document if $c\in\Pi_E(e)$, and zero otherwise. Sums below include only entities with finite positive similarity. With total support $U_c=\sum_e w_{ce}>0$, the Entity score is formulated as:
\begin{equation}
E_c=\operatorname{Norm}(U_c)
\frac{\sum_e w_{ce}\sigma(e)}{U_c}.
\label{eq:entity_score}
\end{equation}
Normalization is applied over passages with positive total support. This weighting reduces the influence of entities that appear in many passages.

\textbf{(3) Relation pooling.} To limit accumulated weak matches, we retain the strongest positive match and pool the remaining support as a residual. We refer to this operation as Relation residual pooling. Let $\mathcal{L}_c$ be the list of finite similarity scores of retrieved relations whose provenance includes passage $c$. If none is positive, we set $H_c=0$. Otherwise, let $s_c^{\max}$ be the largest positive score and compute the score as:
\begin{equation}
H_c=s_c^{\max}+(1-s_c^{\max})
\frac{\sum_{s\in\mathcal{L}_c}\max(0,s)-s_c^{\max}}
{|\mathcal{L}_c|}.
\label{eq:relation_score}
\end{equation}
Nonpositive matches contribute zero to the numerator but remain in the denominator, reducing support from diffuse overlap. We obtain $R_c=\operatorname{Norm}(H_c)$ over passages with $H_c>0$.

\paragraph{Bounded Structural Fusion.}
We rank the union of candidates from all three views, assigning zero to any missing view score. Let $A_c=(E_c+R_c)/2$ denote the combined structural support. The final score adds a bounded structural bonus to Text relevance, which is formulated as:
\begin{equation}
S_c=T_c+\frac{A_c}{1+A_c}.
\label{eq:ratsat}
\end{equation}
Since $A_c\in[0,1]$, the structural bonus lies in $[0,1/2]$ and increases at a decreasing rate. This rule requires no coefficient tuning. \textit{Consensus} returns up to $K_f$ passages with the highest scores and applies the same ranking rule to the original question and all slot-bound subquestions.

\subsection{\textit{Constraint}: Slot-Based Execution}
\label{sec:constraint}

Dependent reasoning steps need earlier answers and supporting context. However, unrestricted information transfer can introduce irrelevant context, and sequential execution can delay independent steps. To address these issues, \textit{Constraint} controls answer transfer, memory sharing, and execution through explicit dependencies.
First, \textit{Consensus} retrieves evidence for $q$. A decomposition LLM uses this evidence to produce an initial summary $I_0$ and a dependency plan. Each step $i$ contains a retrieval-oriented subquestion $q_i$ and an ordered list of direct predecessors $D_i$. The system validates the plan as a directed acyclic graph and determines a topological order.

\paragraph{Slot-Based Answer Transfer.}
The decomposition prompt requires $q_i$ to mark each predecessor's insertion point with a placeholder \texttt{<dep:$j$>} for $j\in D_i$. Once all predecessors finish, \textit{Constraint} replaces each slot with the corresponding short answer to form the bound query $\widetilde{q}_i$. Dependencies specify which answers to transfer, and slots specify where they enter the query.

\paragraph{Dependency-Scoped Memory.}
Each step $i$ produces a short answer $a_i$ and an optional evidence summary $m_i$. Its input memory contains the initial summary followed by summaries from direct predecessors in the declared order of $D_i$, omitting empty summaries. Short answers guide retrieval through query slots, and summaries provide supporting context for reasoning.

\paragraph{Concurrent Dependency Execution.}
All step tasks are created together. Root steps can run immediately, and each remaining step waits for its direct predecessors to finish. Independent branches execute concurrently. Each ready step retrieves fresh evidence for $\widetilde{q}_i$ through \textit{Consensus} and generates its answer and summary from the original question, the bound query, its input memory, and the retrieved passages. After all steps finish, the final model answers $q$ using the plan populated with step answers and the initial summary followed by nonempty step summaries in topological order.

%% file: Sections/4_experiments.tex
\input{Tables/main_results}

\section{Experiments}

\subsection{Experimental Setup}

\paragraph{Tasks and Datasets.}
We conduct experiments on HotpotQA~\citep{yang2018hotpotqa}, 2WikiMultiHopQA~\citep{ho2020twowiki}, and MuSiQue~\citep{trivedi2022musique}.
These datasets cover bridge, comparison, compositional, and multi-hop reasoning questions.
To control cost and ensure comparability, we follow the setting of \citet{chen2026logicrag}, sampling 1,000 questions from the validation split of each dataset and constructing the retrieval corpus with the corresponding supporting passages and distractor passages.
Following prior work~\citep{chen2026logicrag,zhuang2026linearrag}, we adopt two end-to-end question answering metrics: \textit{Str-Acc} and \textit{LLM-Acc}. Specifically, \textit{Str-Acc} denotes the string-based accuracy, which measures whether the prediction contains the reference answer.
\textit{LLM-Acc} denotes the LLM-based accuracy, where a unified LLM judge determines whether the generated answer is semantically consistent with the reference answer~\citep{zheng2023llmjudge}.
Details of all tasks are shown in Appendix~\ref{appendix_data}.

\paragraph{Implementation Details.}
We evaluate \method{} with two LLM backbones, \texttt{GPT-4o-mini}~\citep{openai2024gpt4o} and \texttt{Gemma-4-31B-it}~\citep{gemmateam2026gemma4}. We use \texttt{all-MiniLM-L6-v2}\footnote{\url{https://huggingface.co/sentence-transformers/all-MiniLM-L6-v2}} as the sentence embedding model. By default, we retrieve up to $K_c=16$ objects per view and retain up to $K_f=3$ passages per retrieval. We use greedy decoding throughout inference to support reproducibility. For \textit{LLM-Acc}, we use \texttt{GPT-4o-mini} as the judge across all methods and adopt the same evaluation prompt as \citet{chen2026logicrag}. All experiments conducted in our environment use the same hardware configuration and concurrency limits.

\providecommand{\cmark}{\ding{51}}
\providecommand{\xmark}{\ding{55}}

\begin{figure*}[t]
    \centering
    \begin{minipage}[t]{0.45\textwidth}
        \vspace{0pt}
        \centering
        \includegraphics[width=0.85\linewidth]
            {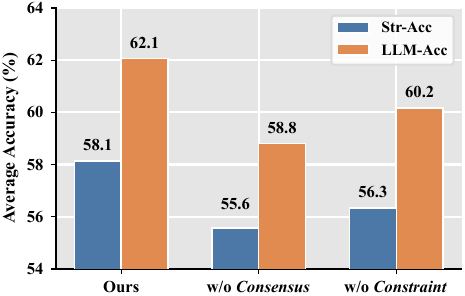}
        \caption{
            \textbf{Ablation results of \textit{Consensus} and \textit{Constraint}.}
            We report the average results of \texttt{GPT-4o-mini}
            across three benchmarks.
        }
        \label{fig:component_ablation}
    \end{minipage}
    \hfill
    \begin{minipage}[t]{0.46\textwidth}
        \vspace{0pt}
        \centering
        \setlength{\tabcolsep}{13pt}
        \resizebox{\linewidth}{!}{%
            \begin{tabular}{@{}ccccc@{}}
                \toprule
                \multicolumn{3}{c}{\textbf{Retrieval Views}}
                & \multicolumn{2}{c}{\textbf{Average Score}} \\
                \cmidrule(lr){1-3}
                \cmidrule(lr){4-5}
                \textbf{Text} & \textbf{Entity} & \textbf{Relation} & \textbf{Str-Acc} & \textbf{LLM-Acc} \\
                \midrule
                \cmark & \xmark & \xmark & 56.1 & 59.3 \\
                \xmark & \cmark & \xmark & 44.6 & 47.6 \\
                \xmark & \xmark & \cmark & 52.0 & 55.3 \\
                \cmark & \cmark & \xmark & 56.6 & 60.2 \\
                \cmark & \xmark & \cmark & 57.4 & 61.2 \\
                \xmark & \cmark & \cmark & 51.3 & 54.7 \\
                \cmark & \cmark & \cmark
                & \textbf{58.1} & \textbf{62.1} \\
                \bottomrule
            \end{tabular}%
        }
        \captionof{table}{
            \textbf{Ablation study on retrieval-view combinations
            in \method{}.}
            We report the average results of \texttt{GPT-4o-mini}
            across three benchmarks.
        }
        \label{tab:view_ablation}
    \end{minipage}
    \vspace{-4pt}
\end{figure*}

\paragraph{Baselines.}
We compare \method{} with a series of baselines:
(1) \textbf{Standard methods}, including Zero-shot and Vanilla RAG~\citep{lewis2020rag,karpukhin2020dpr};
(2) \textbf{Corpus-side methods}, which exploit structured corpus evidence for retrieval, including RAPTOR~\citep{sarthi2024raptor}, G-Retriever~\citep{he2024gretriever}, GraphRAG~\citep{edge2024graphrag}, LightRAG~\citep{guo2025lightrag}, HippoRAG~\citep{gutierrez2024hipporag}, HippoRAG~2~\citep{gutierrez2025hipporag2}, and LinearRAG~\citep{zhuang2026linearrag};
(3) \textbf{Query-side methods}, which adapt retrieval using intermediate reasoning states, including the representative LogicRAG~\citep{chen2026logicrag};
and (4) \textbf{Hybrid methods}, which combine corpus structure with query adaptation, including KGP~\citep{wang2024kgp}, ChainRAG~\citep{zhu2025chainrag}, and Youtu-GraphRAG~\citep{dong2026youtugraphrag}.
Since we mainly follow the experimental setup of LogicRAG, we directly reuse its reported results where available. For the remaining results, we reproduce the baselines using their official implementations. The details of these baselines are shown in Appendix~\ref{appendix_baselines}. 

\subsection{Main Results}

Table~\ref{tab:main_results} presents the main results on several benchmarks, from which we can find that:

\paragraph{\method{} outperforms the other baselines across all benchmarks.}
As seen, \method{} achieves the best results on HotpotQA, 2WikiMultiHopQA, and MuSiQue under both Str-Acc and LLM-Acc metrics.
Compared with Vanilla RAG, \method{} brings clear improvements by using cross-passage connections to retrieve complementary evidence, \textit{e.g.}, +32.1 points on average with \texttt{Gemma-4-31B-it}.
Compared with corpus-side, query-side, and hybrid baselines, \method{} further improves the overall multi-hop QA performance by coordinating noise-aware multi-view ranking with slot-bound answer transfer.
These results demonstrate the effectiveness of coordinating evidence selection and intermediate information transfer for multi-hop QA.

\paragraph{\method{} brings consistent performance gains on both GPT and Gemma backbones.}
The improvements of \method{} are not limited to a specific LLM backbone.
Specifically, with \texttt{GPT-4o-mini}, \method{} achieves the best performance across all three datasets, \textit{e.g.}, +8.0 points on average over the strongest baseline LogicRAG.
Moreover, with \texttt{Gemma-4-31B-it}, \method{} consistently outperforms all multi-hop RAG baselines, \textit{e.g.}, +5.8 points on average over the strongest baseline Youtu-GraphRAG.
These results demonstrate the universality of consensus-driven evidence flow across both LLM backbones.

\paragraph{\method{} is particularly effective in challenging multi-hop scenarios.}
The results also show that \method{} achieves substantial gains on MuSiQue, which requires composing evidence across more complex and less explicit multi-hop chains.
Specifically, \method{} outperforms the strongest baseline by +7.4 and +5.3 points on average with \texttt{GPT-4o-mini} and \texttt{Gemma-4-31B-it}, respectively.
These results demonstrate that \method{} is particularly effective on challenging datasets such as MuSiQue, where evidence must be integrated across longer reasoning chains.

\subsection{Ablation Study}

In this part, we 1) first analyze \textit{Consensus} and \textit{Constraint}, and 2) then examine different combinations of retrieval views.
\texttt{GPT-4o-mini} is used as the default LLM backbone.

\paragraph{Effects of \textit{Consensus} and \textit{Constraint}.}
We evaluate \textit{Consensus} by replacing it with equal-mean fusion (``w/o \textit{Consensus}''), which equally weights the three view scores. For \textit{Constraint}, we disable its core slot-binding mechanism that transfers predecessor answers into dependent retrieval queries (``w/o \textit{Constraint}''). As shown in Figure~\ref{fig:component_ablation}, both variants consistently degrade performance, reducing the average score by 2.9 and 1.8 points, respectively. The former supports the effectiveness of multi-view fusion. The latter demonstrates the benefit of explicit answer transfer through query slots in guiding subsequent retrieval.

\paragraph{Effects of Retrieval-View Combinations.}
Table~\ref{tab:view_ablation} reports the results of different retrieval-view combinations.
Among single-view variants, the Text view achieves the best performance across all three benchmarks.
Adding the Entity or Relation view brings further gains, and combining all three views achieves the best results across all datasets and metrics.
These results show that graph-derived structural signals complement dense textual signals, supporting the effectiveness of \method{}'s consensus-driven multi-view retrieval.

\subsection{Further Analysis}

\input{Tables/fusion_comparison}

\paragraph{Effect of Fusion Strategies.}
To examine how fusion strategies affect answer quality, we compare \textit{Consensus} with Equal-mean, CombMNZ~\citep{fox1994combination}, RRF~\citep{cormack2009rrf}, and RRA~\citep{kolde2012robust}.
Table~\ref{tab:fusion_strategies_main} shows that \textit{Consensus} achieves the highest average score of 60.1 across three benchmarks, outperforming the alternative fusion strategies by 2.9--4.4 points.
\textit{Consensus} retains Text relevance as the base score and adds bounded Entity and Relation support, allowing structural signals to complement textual relevance during passage ranking.
These results demonstrate the superiority of \textit{Consensus}.

\begin{figure}[t]
    \centering
    \includegraphics[width=0.9\linewidth]{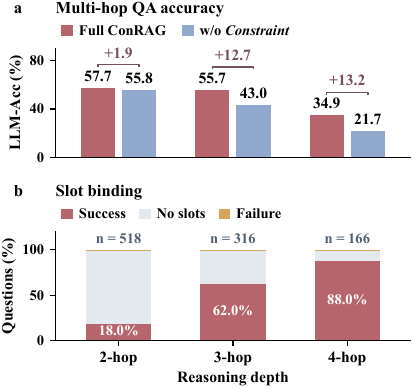}
    \caption{\textbf{Effect of slot binding across reasoning depths on MuSiQue} using \texttt{Gemma-4-31B-it}. (a)~LLM-Acc of Full \method{} and ``w/o \textit{Constraint}''.
(b)~Question proportions by binding outcome.
}
    \label{fig:slot_binding_analysis}
\end{figure}

\paragraph{Effect of Slot Binding.}
To examine how the benefit of slot binding varies with reasoning depth, we compare \method{} with its variant without explicit slot binding (``w/o \textit{Constraint}'') on MuSiQue. Figure~\ref{fig:slot_binding_analysis} shows that the LLM-Acc gains increase from 1.9 points on two-hop questions to 12.7 and 13.2 points on three-hop and four-hop questions, respectively. The proportion of questions with successful slot binding, calculated over all questions in each hop group, also rises from 18.0\% for two-hop questions to 88.0\% for four-hop questions. These results suggest that explicit answer transfer is used more frequently and provides larger benefits in deeper reasoning chains, where bound intermediate answers serve as concrete retrieval targets for dependent steps. Here, successful slot binding means that intermediate answers can be parsed and inserted into the corresponding query slots, but not imply that these answers are correct. We analyze how intermediate answer correctness relates to the benefit of slot binding in Appendix~\ref{appendix_slot_binding_correctness}.

\begin{figure}[t]
    \centering
    \includegraphics[width=0.9\linewidth]{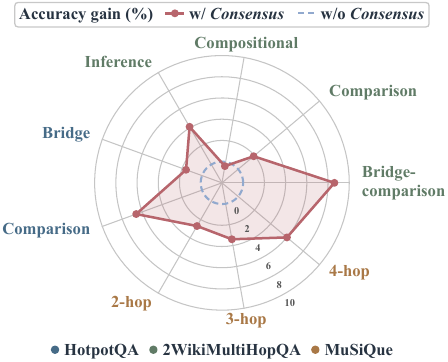}
    \caption{\textbf{Study of \textit{Consensus} across question types and reasoning depths} using \texttt{GPT-4o-mini}.
    Accuracy gains (\%) are the average of Str-Acc and LLM-Acc.
    }
    \label{fig:question_type_analysis}
\end{figure}

\paragraph{Effect of Question Type and Reasoning Depth.}
To examine when \textit{Consensus} is most beneficial, we compare it with equal-mean fusion (``w/o \textit{Consensus}'') across question types and reasoning depths.
Figure~\ref{fig:question_type_analysis} shows Accuracy gains of 8.6 points on 2WikiMultiHopQA bridge-comparison questions and 6.6 points on HotpotQA comparison questions.
On MuSiQue, its benefit grows with reasoning depth, from 2.7 points on two-hop questions to 6.0 points on four-hop questions.
Bridge-comparison questions require combining bridge reasoning with evidence about multiple entities for comparison, while deeper questions require integrating evidence across longer reasoning chains.
These results support \textit{Consensus} as an effective way to improve evidence composition in both branching and deeper reasoning tasks.

\begin{figure}[t]
    \centering
    \includegraphics[width=0.9\linewidth]{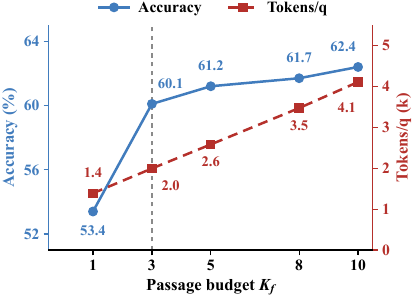}
    \caption{\textbf{Quality-efficiency trade-off with different passage budgets $K_f$} using \texttt{GPT-4o-mini}.
    Accuracy is the average of Str-Acc and LLM-Acc.
    Tokens/q denotes online token consumption per question.
    }
    \label{fig:kf_tradeoff}
\end{figure}

\paragraph{Effect of the Passage Budget $\boldsymbol{K_f}$.}
To examine how the final passage budget affects quality and efficiency, we compare Accuracy and online token consumption across different $K_f$ settings.
As shown in Figure~\ref{fig:kf_tradeoff}, increasing $K_f$ from 1 to 3 improves Accuracy by 6.7 points.
Beyond $K_f=3$, Accuracy gains become more gradual, while online token consumption continues to increase.
Specifically, compared with $K_f=5$ and $K_f=10$, $K_f=3$ reduces online token consumption by 22.8\% and 51.5\%, with Accuracy gaps of just 1.1 and 2.3 points, respectively.
Thus, we use $K_f=3$ as the default setting, as it achieves a favorable quality-efficiency trade-off.

\paragraph{Quality-Efficiency Analysis.}

\begin{figure}[t]
    \centering
    \includegraphics[width=0.9\linewidth]{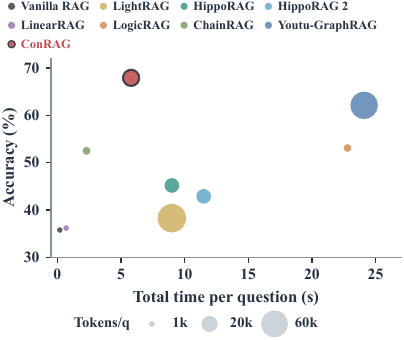}
    \caption{\textbf{Quality-efficiency comparison of different RAG methods across three benchmarks} using \texttt{Gemma-4-31B-it}.
    Accuracy is the average of Str-Acc and LLM-Acc.
    Time and token costs cover the full pipeline and are averaged per question.}
    \label{fig:quality_efficiency_analysis}
\end{figure}

We compare \method{} with other methods in terms of answer accuracy, average runtime per question, and average token consumption per question. Both cost metrics include offline construction costs amortized over the evaluation questions and online inference costs. Figure~\ref{fig:quality_efficiency_analysis} shows that \method{} achieves the highest multi-hop QA accuracy and reduces total runtime and token consumption by 75.9\% and 67.3\%, respectively, compared with the strongest baseline, Youtu-GraphRAG. 
More specifically, during construction, \textit{Connection} extracts entities and 
relations jointly. During inference, \textit{Consensus} retrieves evidence from multiple views in parallel and uses graph results for passage ranking. \textit{Constraint} limits intermediate information transfer to declared dependencies. Together, these mechanisms reduce extraction calls and context size, improving efficiency alongside answer quality.

\paragraph{\ding{43} Note:} Due to space limitations, we introduce more experiments and analyses in Appendix~\ref{sec:appen_experiment}, and more case studies in Appendix~\ref{sec:case_study}.
Please refer to the Appendix for more details.

%% file: Tables/main_results.tex
\begin{table*}[ht]
    \centering
    \setlength{\tabcolsep}{10pt}
    \resizebox{\textwidth}{!}{%
    \begin{tabular}{llcccccccc}
        \toprule
        \multirow{2}{*}{\textbf{Type}}
        & \multirow{2}{*}{\textbf{Method}}
        & \multicolumn{2}{c}{\textbf{HotpotQA}}
        & \multicolumn{2}{c}{\textbf{2WikiMultiHopQA}}
        & \multicolumn{2}{c}{\textbf{MuSiQue}}
        & \multicolumn{2}{c}{\textbf{Overall}} \\
        \cmidrule(lr){3-4}
        \cmidrule(lr){5-6}
        \cmidrule(lr){7-8}
        \cmidrule(lr){9-10}
        &
        & \textbf{Str-Acc} & \textbf{LLM-Acc}
        & \textbf{Str-Acc} & \textbf{LLM-Acc}
        & \textbf{Str-Acc} & \textbf{LLM-Acc}
        & \textbf{Avg.} & $\boldsymbol{\Delta}\uparrow$ \\
        \midrule

        \rowcolor{gray!15}
        \multicolumn{10}{l}{\textit{\textbf{GPT-4o-mini} is used as the LLM backbone}} \\

        \multirow{2}{*}{Standard}
        & Zero-shot$^\dag$
        & 38.7 & 36.3 & 26.4 & 24.3 & 17.6 & 14.0
        & 26.2 & $-11.3$ \\
        & Vanilla RAG$^\dag$
        & 43.2 & 53.1 & 43.0 & 42.0 & 20.3 & 23.6
        & 37.5 & -- \\
        \midrule

        \multirow{7}{*}{Corpus-side}
        & RAPTOR$^\dag$
        & 48.1 & 57.8 & 47.7 & 45.9 & 25.2 & 29.1
        & 42.3 & $+4.8\phantom{0}$ \\
        & G-Retriever$^\dag$
        & 28.5 & 40.9 & 26.7 & 35.7 & 9.1 & 15.6
        & 26.1 & $-11.4$ \\
        & GraphRAG$^\dag$
        & 39.6 & 45.2 & 46.3 & 43.3 & 16.5 & 23.1
        & 35.7 & $-1.8\phantom{0}$ \\
        & LightRAG$^\dag$
        & 47.8 & 57.7 & 43.1 & 36.3 & 18.1 & 19.4
        & 37.1 & $-0.4\phantom{0}$ \\
        & HippoRAG$^\dag$
        & 53.7 & 55.6 & 47.7 & 47.2 & 24.9 & 30.1
        & 43.2 & $+5.7\phantom{0}$ \\
        & HippoRAG~2$^\dag$
        & 56.7 & 61.9 & 50.0 & 47.1 & 27.0 & 32.6
        & 45.9 & $+8.4\phantom{0}$ \\
        & LinearRAG
        & 50.1 & 49.1 & 44.7 & 45.7 & 17.4 & 17.9
        & 37.5 & $\phantom{+}0.0\phantom{0}$ \\
        \midrule

        \multirow{1}{*}{Query-side}
        & LogicRAG$^\dag$
        & 54.8 & 62.6 & 64.7 & 62.5 & 30.4 & 37.5
        & 52.1 & $+14.6$ \\
        \midrule

        \multirow{3}{*}{Hybrid}
        & KGP$^\dag$
        & 46.4 & 57.1 & 47.5 & 43.7 & 23.3 & 27.5
        & 40.9 & $+3.4\phantom{0}$ \\
        & ChainRAG
        & 55.8 & 61.5 & 57.4 & 57.2 & 30.4 & 34.7
        & 49.5 & $+12.0$ \\
        & Youtu-GraphRAG
        & 54.5 & 56.1 & 53.7 & 56.6 & 29.9 & 30.2
        & 46.8 & $+9.3\phantom{0}$ \\
        \midrule

        \rowcolor[RGB]{233,246,255}
        Ours
        & \textbf{\method{}}
        & \textbf{61.3} & \textbf{68.3}
        & \textbf{74.2} & \textbf{74.1}
        & \textbf{38.9} & \textbf{43.8}
        & \textbf{60.1} & \makebox[0pt][l]{$\mathbf{+22.6}$}\phantom{$+22.6$} \\
        \midrule

        \rowcolor{gray!15}
        \multicolumn{10}{l}{\textit{\textbf{Gemma-4-31B-it} is used as the LLM backbone}} \\

        \multirow{2}{*}{Standard}
        & Zero-shot
        & 31.5 & 30.6 & 31.8 & 31.4 & 9.3 & 9.3
        & 24.0 & $-11.8$ \\
        & Vanilla RAG
        & 47.6 & 48.3 & 37.5 & 36.9 & 22.0 & 22.7
        & 35.8 & -- \\
        \midrule

        \multirow{4}{*}{Corpus-side}
        & LightRAG
        & 51.4 & 55.7 & 39.2 & 33.7 & 22.7 & 26.8
        & 38.3 & $+2.5\phantom{0}$ \\
        & HippoRAG
        & 54.1 & 55.3 & 52.9 & 55.6 & 25.4 & 27.6
        & 45.2 & $+9.4\phantom{0}$ \\
        & HippoRAG~2
        & 54.1 & 55.9 & 46.1 & 47.5 & 25.7 & 27.9
        & 42.9 & $+7.1\phantom{0}$ \\
        & LinearRAG
        & 45.7 & 46.1 & 43.0 & 44.2 & 18.3 & 19.7
        & 36.2 & $+0.4\phantom{0}$ \\
        \midrule

        \multirow{1}{*}{Query-side}
        & LogicRAG
        & 60.7 & 62.4 & 60.7 & 65.7 & 32.1 & 37.2
        & 53.1 & $+17.3$ \\
        \midrule

        \multirow{2}{*}{Hybrid}
        & ChainRAG
        & 58.4 & 62.5 & 55.1 & 54.4 & 39.6 & 44.8
        & 52.5 & $+16.7$ \\
        & Youtu-GraphRAG
        & 67.4 & 69.1 & 69.5 & 74.7 & 43.8 & 48.3
        & 62.1 & $+26.3$ \\
        \midrule

        \rowcolor[RGB]{233,246,255}
        Ours
        & \textbf{\method{}}
        & \textbf{68.4} & \textbf{72.4}
        & \textbf{82.9} & \textbf{80.8}
        & \textbf{49.4} & \textbf{53.3}
        & \textbf{67.9} & \makebox[0pt][l]{$\mathbf{+32.1}$}\phantom{$+32.1$} \\
        \bottomrule
    \end{tabular}%
    }
    \caption{\textbf{Performance comparison (\%) on three multi-hop QA benchmarks.}
    ``Avg.'' denotes the average results and ``$\Delta\uparrow$'' refers to the gains against Vanilla RAG. ``$\dag$'' indicates the results collected from~\citet{chen2026logicrag}.
    }
    \label{tab:main_results}
\end{table*}

%% file: Tables/fusion_comparison.tex
\begin{table}[t]
    \centering
    \setlength{\tabcolsep}{18pt}
    \resizebox{0.48\textwidth}{!}{%
    \begin{tabular}{@{}lccc@{}}
        \toprule
        \multirow{2}{*}{\textbf{Fusion}}
        & \multicolumn{3}{c}{\textbf{Average Score}} \\
        \cmidrule(lr){2-4}
        & \textbf{Str-Acc} & \textbf{LLM-Acc} & \textbf{Avg.} \\
        \midrule
        Equal-mean & 55.6 & 58.8 & 57.2 \\
        CombMNZ    & 55.1 & 58.9 & 57.0 \\
        RRF        & 53.8 & 57.5 & 55.7 \\
        RRA        & 55.4 & 58.8 & 57.1 \\
        \midrule
        \textbf{\textit{Consensus}}
        & \textbf{58.1} & \textbf{62.1} & \textbf{60.1} \\
        \bottomrule
    \end{tabular}%
    }
    \caption{\textbf{Performance comparison (\%) of fusion strategies in \textit{Consensus}.}
    Here, we report the average results of \texttt{GPT-4o-mini} across three benchmarks.}
    \label{tab:fusion_strategies_main}
\end{table}

%% file: Sections/5_conclusion.tex
\section{Conclusion}

In this paper, we propose \method{}, a consensus-driven evidence flow framework for complex multi-hop QA. In short, \method{} connects evidence across passages and integrates Text, Entity, and Relation signals through multi-view ranking with bounded structural support. It also uses slot-bound answer transfer and dependency-scoped memory to guide retrieval and reasoning across dependent steps. Together, these mechanisms coordinate evidence selection and intermediate information transfer. Experiments on three widely used multi-hop QA benchmarks demonstrate that \method{} consistently outperforms existing RAG baselines and improves efficiency.

%% file: Sections/6_limitations.tex
\section*{Limitations}

Our work has several potential limitations.
First, due to limited computational budget, we only evaluate \method{} on three English multi-hop QA benchmarks with two LLM backbones. Extending the experiments to more languages, larger corpora, and more LLM backbones would further assess the generalizability of our findings.
Second, \method{} relies on offline provenance-preserving graph construction and multi-view index building. Although this design keeps the online retrieval cost controllable, graph construction quality may still be affected by information extraction errors, noisy relations, and entity normalization mistakes. How to build and incrementally update the provenance-preserving graph more efficiently in large-scale or frequently changing corpora remains under-explored.
Lastly, slot-bound execution in \method{} depends on the quality of intermediate answers. When predecessor subquestions produce incorrect or incomplete answers, the slot-bound queries may introduce misleading constraints for subsequent retrieval steps. We believe that incorporating intermediate answer verification or selective re-planning represents a promising direction for future work.

%% file: Sections/7_ethics_statement.tex
\section*{Ethics Statement}

We take ethical considerations very seriously and strictly adhere to the ACL Ethics Policy.
This paper proposes a retrieval framework for complex multi-hop question answering.
It aims to improve evidence retrieval and answer generation over provided corpora, rather than encouraging models to learn private or sensitive knowledge.
All datasets used in this paper are publicly available and have been widely adopted by researchers.
The constructed evidence-grounded knowledge graph is used only for retrieval and ranking.
Therefore, we do not anticipate any additional ethical concerns.

%% file: Sections/8_appendix.tex
\appendix
\section{Appendix}
\label{sec:appendix}

\subsection{Details of Tasks and Datasets}
\label{appendix_data}

We conduct experiments on three widely used multi-hop QA benchmarks: HotpotQA~\citep{yang2018hotpotqa}, 2WikiMultiHopQA~\citep{ho2020twowiki}, and MuSiQue~\citep{trivedi2022musique}. Following \citet{chen2026logicrag}, we evaluate on 1,000 questions from the validation split of each benchmark and construct the retrieval corpus from the corresponding supporting and distractor passages.
Notably, the HotpotQA corpus size reported in the LogicRAG paper differs from that of the dataset released in its official repository. We use the publicly released version\footnote{\url{https://github.com/chensyCN/LogicRAG}} and rerun LogicRAG to check comparability. Our reproduced results closely match the reported scores, with Str-Acc of 54.1 versus 54.8 and LLM-Acc of 62.0 versus 62.6. These results suggest that the discrepancy has a limited effect on the comparison. Table~\ref{tab:dataset_statistics} summarizes the dataset statistics. We briefly describe each benchmark below.

\paragraph{HotpotQA.}
HotpotQA~\citep{yang2018hotpotqa} is a Wikipedia-based multi-hop question answering benchmark.
It mainly contains two types of questions, bridge and comparison questions, which require systems to retrieve and combine supporting evidence across multiple passages.

\paragraph{2WikiMultiHopQA.}
2WikiMultiHopQA~\citep{ho2020twowiki} is constructed from Wikipedia and Wikidata.
It contains four types of multi-hop questions that require comparison, composition, inference, and bridge comparison over entities.
These questions are grounded in entity relations, making the dataset suitable for evaluating relation-aware retrieval and multi-step reasoning.

\paragraph{MuSiQue.}
MuSiQue~\citep{trivedi2022musique} is a challenging multi-hop QA benchmark designed to reduce shortcut reasoning.
It contains three types of compositional questions with different reasoning depths.
Compared with HotpotQA and 2WikiMultiHopQA, its questions usually involve less explicit but more compositional evidence chains.

\input{Tables/dataset_statistics}

\subsection{Details of Compared Methods}
\label{appendix_baselines}

As listed in Table~\ref{tab:main_results}, we compare \method{} with representative baselines from the four categories: standard baselines, corpus-side methods, query-side methods, and hybrid methods. The grouping reflects the primary mechanism emphasized in our comparison, and it is organizational rather than mutually exclusive. In the main experiments, we directly reuse the \texttt{GPT-4o-mini} results reported by \citet{chen2026logicrag} for all baselines except LinearRAG, ChainRAG, and Youtu-GraphRAG.
All other results are reproduced by using the official implementations of corresponding methods.

\paragraph{Zero-shot.}
Zero-shot directly prompts the backbone LLM with the original question and generates an answer without external retrieval.
It serves as a reference for the parametric knowledge and reasoning ability of each backbone.

\paragraph{Vanilla RAG.}
Vanilla RAG retrieves text chunks according to query--chunk similarity and feeds the top-ranked chunks into the LLM for answer generation.
It represents the standard chunk-based retrieval paradigm without corpus-side graph structures or query-side decomposition.

\paragraph{RAPTOR.}
RAPTOR~\citep{sarthi2024raptor} recursively clusters and summarizes text segments to construct a hierarchical corpus representation.
The retrieval space of RAPTOR contains both original text segments and summary nodes at multiple levels of abstraction.

\paragraph{G-Retriever.}
G-Retriever~\citep{he2024gretriever} operates over textual graphs by assigning query-dependent relevance prizes to nodes and edges.
It formulates connected-subgraph selection as a Prize-Collecting Steiner Tree problem and supplies both textualized and learned graph representations to the LLM.

\paragraph{GraphRAG.}
GraphRAG~\citep{edge2024graphrag} extracts entities and relations from the corpus, identifies hierarchical graph communities, and constructs community reports.
At query time, it aggregates information from these reports to provide corpus-level context for answer generation.

\paragraph{LightRAG.}
LightRAG~\citep{guo2025lightrag} constructs an entity--relation graph while retaining links to the underlying text.
Its dual-level retrieval combines local entity cues with global relation or thematic cues.

\paragraph{HippoRAG.}
HippoRAG~\citep{gutierrez2024hipporag}, inspired by human long-term memory, constructs a schemaless phrase graph from OpenIE triples and embedding-based synonym links.
It initializes Personalized PageRank from query-related concepts and aggregates graph-node scores back to their source passages.

\paragraph{HippoRAG~2.}
HippoRAG~2~\citep{gutierrez2025hipporag2} extends HippoRAG by incorporating passage nodes into the phrase graph.
It initializes retrieval from query-matched triples and passages, filters triple seeds with an LLM, and ranks passage nodes through Personalized PageRank.

\paragraph{LinearRAG.}
LinearRAG~\citep{zhuang2026linearrag} constructs a relation-free entity--sentence--passage Tri-Graph without LLM-based relation extraction.
It combines query-conditioned semantic propagation for bridge-entity discovery with graph-based passage ranking.

\paragraph{LogicRAG.}
LogicRAG~\citep{chen2026logicrag} constructs an inference-time dependency DAG over query-derived subproblems rather than pre-building a corpus graph.
It executes subproblems in topological order and conditions later retrieval steps on preceding results.

\paragraph{KGP.}
KGP~\citep{wang2024kgp} represents passages and document structures as nodes connected through lexical, semantic, entity-based, or structural relations.
An LLM generates a description of the next required evidence and grounds it to neighboring graph nodes during iterative traversal.

\paragraph{ChainRAG.}
ChainRAG~\citep{zhu2025chainrag} first constructs a sentence graph using entity co-occurrence, semantic similarity, and document adjacency.
Then, ChainRAG retrieves evidence for successive subquestions and rewrites dependent queries using previously generated answers to recover missing entities.

\paragraph{Youtu-GraphRAG.}
Youtu-GraphRAG~\citep{dong2026youtugraphrag} uses a shared schema to guide graph construction, hierarchical knowledge organization, and query interpretation.
Its retrieval agent decomposes complex questions and routes subqueries across entity-, triple-, community-, and path-based retrieval operations.
To reduce the overhead of repeatedly scanning the node mapping for each candidate during type-filtered retrieval, our reproduction uses direct index lookups while preserving original retrieval logic.

\input{Tables/consensus_ablation_analysis}

\subsection{More Detailed Procedures in \method{}}
\label{appendix_algorithms}

Algorithms~\ref{alg:connection}--\ref{alg:constraint} instantiate the three key operations of \method{} in Section~\ref{sec:method}.
Specifically, \textit{Connection} constructs the provenance-preserving three-view interface once offline. \textit{Consensus} ranks passage candidates over this interface, and \textit{Constraint} repeatedly reuses the same retriever through slot-bound queries during inference.

\begin{algorithm}[t]
\small
\caption{\textit{Connection} in \method{}}
\label{alg:connection}
\begin{algorithmic}[1]
\Require Passages $\mathcal{P}=\{p_c\}_{c=1}^{n}$, graph extractor, encoder, worker bound $W$
\Ensure Global graph $\mathcal{G}$, provenance maps $\Pi_E,\Pi_R$, indices $\mathcal{I}_T,\mathcal{I}_E,\mathcal{I}_R$
\State Launch $G_c\gets\Call{ExtractPassageGraph}{p_c}$ for all $c$ with at most $W$ workers
\State $\mathcal{G}\gets\Call{MultiDiGraph}{}$; initialize empty maps $\Pi_E,\Pi_R$ with default $\varnothing$
\ForAll{completed $(c,G_c)$}
    \ForAll{$e\in\mathcal{E}_c$}
        \State $e^\star\gets\Call{MergeEntity}{\mathcal{G},e,\kappa_E(e)}$
        \State $\Pi_E(e^\star)\gets\Pi_E(e^\star)\cup\{c\}$
    \EndFor
    \ForAll{$\rho\in\mathcal{R}_c$}
        \State $\rho^\star\gets\Call{MergeRelation}{\mathcal{G},\rho,\kappa_R(\rho)}$
        \State $\Pi_R(\rho^\star)\gets\Pi_R(\rho^\star)\cup\{c\}$
    \EndFor
\EndFor
\State Serialize Text, Entity, and Relation documents as in Section~\ref{sec:connection}
\State Encode the three document views and $\ell_2$-normalize their embedding matrices
\State Build exact inner-product indices $\mathcal{I}_T,\mathcal{I}_E,\mathcal{I}_R$
\State \Return $\mathcal{G},\Pi_E,\Pi_R,\mathcal{I}_T,\mathcal{I}_E,\mathcal{I}_R$
\end{algorithmic}
\end{algorithm}

\paragraph{Merge-Key Normalization.}
In Algorithm~\ref{alg:connection}, $G_c$ denotes the graph extracted from passage $p_c$, with entity set $\mathcal{E}_c$ and relation set $\mathcal{R}_c$. The functions $\kappa_E$ and $\kappa_R$ return the entity and relation merge keys described in Section~\ref{sec:connection}. The indices $\mathcal{I}_T$, $\mathcal{I}_E$, and $\mathcal{I}_R$ correspond to the Text, Entity, and Relation views, respectively.
Before constructing $\kappa_E$ and $\kappa_R$, \method{} applies Unicode NFKC normalization, standardizes dash and quotation characters, collapses whitespace, removes surrounding punctuation, and preserves case.
The relation key combines the normalized relation label with the normalized name--type identities of its ordered source and target entities.

\paragraph{Score Normalization.}
In Algorithm~\ref{alg:consensus}, $\operatorname{dom}$ denotes the passage IDs stored in a sparse map. The maps $T$, $E$, and $R$ contain the corresponding view scores. The accumulator $V_c=\sum_e w_{ce}\sigma(e)$ stores specificity-weighted Entity support, so $V_c/U_c$ is the weighted specificity in Equation~\ref{eq:entity_score}. The procedures \textsc{IDFWeight} and \textsc{RelationPool} implement Equations~\ref{eq:entity_specificity} and~\ref{eq:relation_score}, respectively.
\textit{Consensus} removes nonfinite entries before normalizing each sparse, view-specific passage map.
An empty map remains empty, and every candidate in a nonempty constant map receives one.
Passages absent from a view receive zero during fusion.
For Relation evidence, \textit{Consensus} retains only candidates with $H_c>0$ before normalizing the resulting map.

\paragraph{Plan Normalization.}
In Algorithm~\ref{alg:constraint}, each step is written as $s_i=(i,q_i,D_i)$. The validated plan is $\mathcal{S}=(s_{\pi_1},\ldots,s_{\pi_L})$, where $L$ is the number of steps and $\pi$ gives a topological ordering of their IDs.
The planning prompt requires zero-based contiguous IDs, earlier-step dependencies, and consistency between declared dependencies and slots.
The runtime validates unique contiguous integer IDs, nonempty subquestions, integer dependency lists, existing dependency IDs, and DAG acyclicity before returning a topological order.
If decomposition or normalization yields no valid plan, \textit{Constraint} instantiates the single-step fallback $s_0=(0,q,[\ ])$.

\paragraph{Execution Notation.}
In Algorithm~\ref{alg:constraint}, $C_{\mathrm{init}}$ and $C_i$ denote passages retrieved for the original question and bound subquestion, respectively. The procedure \textsc{BindSlots} replaces each \texttt{<dep:$j$>} placeholder with answer $a_j$. The symbol $\varnothing$ denotes an empty summary, $[\ ]$ an empty sequence, and $\mathbin{\Vert}$ concatenation. Predecessor summaries follow the declared order of $D_i$. Each $\operatorname{Event}_i$ signals completion of step $i$.

\paragraph{Execution Fallbacks.}
A failed passage extraction contributes an empty passage-level graph.
A failed plan step records \texttt{Information not found}, emits no evidence summary, and signals completion so dependent steps can proceed.
A final-generation failure returns \texttt{Information not found}.
Final synthesis consumes the original question, rendered plan answers, and ordered evidence-summary memory.

\begin{algorithm}[t]
\small
\caption{\textit{Consensus} in \method{}}
\label{alg:consensus}
\begin{algorithmic}[1]
\Require Query $q$, encoder, corpus size $n$, indices $\mathcal{I}_T,\mathcal{I}_E,\mathcal{I}_R$, provenance $\Pi_E,\Pi_R$, per-view candidate budget $K_c$, final passage budget $K_f$
\Ensure Up to $K_f$ ranked passages
\State $\mathbf z_q\gets\Call{EncodeQuery}{q}$; $\mathbf z_q\gets\Call{L2Normalize}{\mathbf z_q}$
\State Using $\mathbf z_q$, retrieve up to $K_c$ hits from each index in parallel as $\mathcal{H}_T,\mathcal{H}_E,\mathcal{H}_R$
\State $T\gets\Call{Norm}{\{c\mapsto s\mid(c,s)\in\mathcal{H}_T\}}$
\State Initialize empty sparse maps $U,V$ with default $0$, and $\mathcal{L}$ with default $[\ ]$
\ForAll{$(e,s)\in\mathcal{H}_E$ with finite $s>0$}
    \State $\sigma(e)\gets\Call{IDFWeight}{|\Pi_E(e)|,n}$
    \ForAll{$c\in\Pi_E(e)$}
        \State $U_c\gets U_c+s$; $V_c\gets V_c+s\sigma(e)$
    \EndFor
\EndFor
\State $\mathcal{C}_E\gets\operatorname{dom}(U)$; $\widehat U\gets\Call{Norm}{U}$
\State $E\gets\{c\mapsto\widehat U_c(V_c/U_c)\mid c\in\mathcal{C}_E\}$
\ForAll{$(\rho,s)\in\mathcal{H}_R$ with finite $s$}
    \ForAll{$c\in\Pi_R(\rho)$}
        \State Append $s$ to $\mathcal{L}_c$
    \EndFor
\EndFor
\State $H\gets\{c\mapsto\Call{RelationPool}{\mathcal{L}_c}\mid c\in\operatorname{dom}(\mathcal{L})\}$
\State $\mathcal{C}_R\gets\{c\in\operatorname{dom}(H)\mid H_c>0\}$
\State $R\gets\Call{Norm}{\{c\mapsto H_c\mid c\in\mathcal{C}_R\}}$
\State $\mathcal{C}\gets\operatorname{dom}(T)\cup\operatorname{dom}(E)\cup\operatorname{dom}(R)$
\ForAll{$c\in\mathcal{C}$}
    \State $A_c\gets(E_c+R_c)/2$, with missing view scores set to $0$
    \State $S_c\gets T_c+A_c/(1+A_c)$, with missing $T_c$ set to $0$
\EndFor
\State \Return up to $K_f$ passages in $\mathcal{C}$ in descending order of $S_c$
\end{algorithmic}
\end{algorithm}

\begin{algorithm}[t]
\small
\caption{\textit{Constraint} in \method{}}
\label{alg:constraint}
\begin{algorithmic}[1]
\Require Question $q$, \textit{Consensus} retriever, planner, StepLLM, and FinalLLM
\Ensure Final answer $\hat{y}$
\State $C_{\mathrm{init}}\gets\Call{ConsensusRetrieve}{q}$
\State $(I_0,\mathcal{S}_{\mathrm{raw}})\gets\Call{Planner}{q,C_{\mathrm{init}}}$
\State $\mathcal{S}\gets\Call{NormalizePlan}{\mathcal{S}_{\mathrm{raw}}}$
\If{$\mathcal{S}$ is empty}
    \State $\mathcal{S}\gets(s_0)$, where $s_0=(0,q,[\ ])$
\EndIf
\State $L\gets|\mathcal{S}|$; write $\mathcal{S}=(s_{\pi_1},\ldots,s_{\pi_L})$
\State $\mathcal{M}_{\mathrm{base}}\gets[I_0]$ if $I_0\neq\varnothing$, otherwise $[\ ]$
\State Initialize $a_i\gets\texttt{Information not found}$, $m_i\gets\varnothing$, and an unset $\operatorname{Event}_i$ for every step
\ForAll{$k\in\{1,\ldots,L\}$ \textbf{in parallel}}
    \State $i\gets\pi_k$
    \State \textbf{try:}
    \State Wait until $\operatorname{Event}_j$ is set for every $j\in D_i$
    \State $\widetilde q_i\gets\Call{BindSlots}{q_i,\{(j,a_j):j\in D_i\}}$
    \State $\mathcal{M}_i\gets\mathcal{M}_{\mathrm{base}}\mathbin{\Vert}[m_j:j\in D_i\land m_j\neq\varnothing]$
    \State $C_i\gets\Call{ConsensusRetrieve}{\widetilde q_i}$
    \State $(a_i,m_i)\gets\Call{StepLLM}{q,\widetilde q_i,\mathcal{M}_i,C_i}$
    \State \textbf{on failure:} $a_i\gets\texttt{Information not found}$; $m_i\gets\varnothing$
    \State \textbf{finally:} set $\operatorname{Event}_i$
\EndFor
\State Wait for all step tasks to finish
\State $\mathcal{M}_{\mathrm{steps}}\gets[m_{\pi_k}:1\leq k\leq L\land m_{\pi_k}\neq\varnothing]$
\State $\mathcal{M}_{\mathrm{final}}\gets\mathcal{M}_{\mathrm{base}}\mathbin{\Vert}\mathcal{M}_{\mathrm{steps}}$
\State $E_P\gets\Call{RenderPlan}{\mathcal{S},\{a_i\}}$
\State $E_M\gets\Call{RenderMemory}{\mathcal{M}_{\mathrm{final}}}$
\State \Return $\Call{FinalLLM}{q,E_P\mathbin{\Vert}E_M}$
\end{algorithmic}
\end{algorithm}

\subsection{Additional Analysis and Full Results}
\label{sec:appen_experiment}

Here, we further analyze parameter sensitivity and additional answer quality metrics, and then report the full results of experiments in the main paper.

\paragraph{Effects of \textit{Consensus} Scoring Mechanisms.}
Here, we conduct ablation studies to examine the effects of Entity specificity and Relation residual pooling in \textit{Consensus}.
We remove Entity specificity or replace Relation residual pooling with maximum pooling, while keeping \textit{Connection} and \textit{Constraint} unchanged.
As shown in Table~\ref{tab:consensus_ablation}, removing either mechanism consistently degrades performance on all three benchmarks.
Specifically, the overall average drops by 0.5 and 0.9 points, respectively.
The largest decrease occurs when replacing Relation residual pooling on 2WikiMultiHopQA, where LLM-Acc drops by 2.2 points.
These results demonstrate the effectiveness of both scoring mechanisms in \textit{Consensus}.
The improvement over maximum pooling further shows the benefit of accounting for weaker relation matches when evaluating Relation evidence.

\begin{figure}[t]
    \centering
    \includegraphics[width=\linewidth]{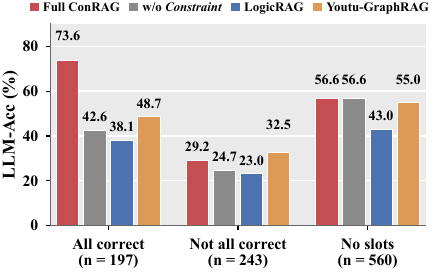}
    \caption{\textbf{Effect of intermediate answer correctness on MuSiQue} using \texttt{Gemma-4-31B-it}. We report final-answer LLM-Acc on the same question groups, defined by slot use and intermediate answer correctness in \method{}. $n$ denotes the number of questions in each group.}
    \label{fig:slot_binding_correctness}
\end{figure}

\paragraph{Effect of Intermediate Answer Correctness.}
\label{appendix_slot_binding_correctness}
To examine how intermediate answer correctness relates to the benefit of slot binding, we compare answer accuracy across correctness groups on MuSiQue using \texttt{Gemma-4-31B-it}. We evaluate intermediate answers with the same \texttt{GPT-4o-mini} judge used in the main experiments, using references derived from MuSiQue's gold question decomposition and supporting passages. First, we divide questions that use slots into ``All correct'' and ``Not all correct'' groups based on whether all non-terminal answers produced by \method{} are judged correct. Within each group, we compare \method{} with its ``w/o \textit{Constraint}'' variant and two strong baselines, LogicRAG and Youtu-GraphRAG, on the same questions. Figure~\ref{fig:slot_binding_correctness} shows that \method{} achieves 73.6\% LLM-Acc in the ``All correct'' group, outperforming ``w/o \textit{Constraint}'' by 31.0 points. In the ``Not all correct'' group, LLM-Acc ranges from 23.0\% to 32.5\% across methods, and the gain over ``w/o \textit{Constraint}'' is 4.5 points. Both variants perform equally on questions without slots. These results suggest that slot binding supports information transfer across reasoning steps, with larger benefits when intermediate answers are correct. However, its gains are smaller when intermediate errors occur, highlighting its dependence on answer correctness. We acknowledge this limitation in the Limitations and identify intermediate answer verification as a direction for future work.

\input{Tables/kc_sensitivity}
\input{Tables/additional_metrics}

\paragraph{Parameter Analysis.}
Here, we examine the influence of the candidate budget $K_c$ with $K_f=3$ and \texttt{GPT-4o-mini} fixed.
As shown in Table~\ref{tab:kc_sensitivity}, \method{} performs consistently across $K_c\in\{8,16,24,32\}$.
The overall Str-Acc ranges from 57.6 to 58.1, while LLM-Acc ranges from 61.5 to 62.1.
These results indicate that \method{} is not sensitive to $K_c$ within the tested range.
More specifically, $K_c=16$ achieves the highest overall Str-Acc and matches $K_c=24$ for the highest overall LLM-Acc.
We therefore use $K_c=16$ as the default setting in our experiments.

\paragraph{Additional Answer Quality Analysis.}
To further evaluate answer quality, we report exact match (EM) and token-level F1 results with \texttt{Gemma-4-31B-it}.
As shown in Table~\ref{tab:additional_metrics}, \method{} achieves the best EM and F1 scores on all three benchmarks.
Specifically, its average across both metrics and all three benchmarks reaches 62.9, compared with 61.8 for Youtu-GraphRAG and 35.5 for Vanilla RAG.
The ranking is consistent with the main results, although the margin over Youtu-GraphRAG is smaller under EM and F1. This further demonstrates the effectiveness of \method{} for multi-hop QA.

\input{Tables/stage_wise_efficiency}

\paragraph{Stage-Wise Efficiency Analysis.}
To examine how costs are distributed across stages, we separate offline construction from online inference on 2WikiMultiHopQA using \texttt{Gemma-4-31B-it}.
As shown in Table~\ref{tab:stage_wise_efficiency}, \method{} incurs an amortized construction cost of 1.6 seconds and 11,951.7 tokens per question.
Both are lower than LightRAG's corresponding costs of 4.7 seconds and 40,025.4 tokens.
After construction, \method{} requires 2.7 seconds and 2,362.1 tokens per question for online inference.
Compared with LogicRAG and Youtu-GraphRAG, \method{} reduces online latency by 85.6\% and 80.9\%, and online token consumption by 22.5\% and 75.6\%, respectively.
Offline construction accounts for 83.5\% of \method{}'s total token consumption under this evaluation workload.
Since the resulting graph and indices are reused, subsequent queries over the same corpus can amortize this initial investment without repeating graph extraction.

\paragraph{Full Results.}
We report the full results of experiments in the main paper.
Table~\ref{tab:component_ablation} and Table~\ref{tab:view_ablation_full} show the component and retrieval-view ablations, respectively.
Table~\ref{tab:fusion_comparison} shows the detailed results of different fusion strategies.
Table~\ref{tab:slot_binding_analysis} shows the detailed results of slot binding on MuSiQue.
Table~\ref{tab:question_type_analysis} shows the detailed results across question types and reasoning depths.
Table~\ref{tab:top-k_analysis} shows the detailed results with different passage budgets $K_f$.
Table~\ref{tab:efficiency_analysis} reports the total time and token consumption per question on each benchmark.

\input{Tables/component_ablation}
\input{Tables/view_ablation_full}
\input{Tables/full_fusion_comparison}
\input{Tables/slot_binding_analysis}
\input{Tables/question_type_analysis}
\input{Tables/top-k_analysis}
\input{Tables/efficiency_analysis}

\subsection{Case Study}
\label{sec:case_study}

To better understand \textit{Consensus} and \textit{Constraint}, we present a four-hop MuSiQue example in Table~\ref{tab:case_study} and Table~\ref{tab:case_study_constraint}.
The question requires connecting \textit{Changes Two}, Charles Mingus, Arizona, and Phoenix to identify the race winner Mario Andretti.
As shown in Table~\ref{tab:case_study}, both Text and Equal-mean rank the answer-bearing passage P4 fourth, whereas \textit{Consensus} ranks it third.
Thus, \textit{Consensus} retrieves this passage within the same top-three budget, showing its effectiveness in selecting complementary evidence.

We further compare \method{} with the variant without explicit slot binding in Table~\ref{tab:case_study_constraint}.
Both executions retrieve the same initial passages and correctly identify Charles Mingus at step 0.
Without slot binding, later queries retain indirect descriptions of the performer and location, and the execution incorrectly selects Oklahoma City.
In contrast, \method{} passes Charles Mingus, Arizona, and Phoenix into successive queries, retrieves P2, P3, and P4, and correctly answers Mario Andretti.
This example shows how explicitly passing intermediate answers guides subsequent retrieval to complete the multi-hop reasoning chain.

\input{Tables/case_study}

\subsection{Prompt Details}
\label{appendix_prompts}

Here, we present the detailed prompts used by \method{}.
Specifically, we present the prompts and schema for knowledge graph extraction in Figures~\ref{fig:prompt_knowledge_graph_extraction} and~\ref{fig:knowledge_graph_schema}, question decomposition in Figure~\ref{fig:prompt_question_decomposition}, subquestion answering in Figure~\ref{fig:prompt_subquestion_answering}, and final answer synthesis in Figure~\ref{fig:prompt_final_answer_synthesis}.
In particular, the knowledge graph extraction prompt uses the schema to extract explicit entities, attributes, and directed relations from each passage for \textit{Connection}.
The question decomposition prompt uses the evidence initially retrieved by \textit{Consensus} to organize subquestions and their dependencies for \textit{Constraint}.
The subquestion answering prompt produces a short answer and an evidence summary using retrieved passages and dependency-scoped memory.
The answer synthesis prompt combines the answered plan and ordered evidence summaries to answer the original question.

\input{Figures/prompt_figures}

\subsection{Use of AI Assistants}

During the writing of this paper, we used proprietary LLMs only as general-purpose writing assistants.
They were used to polish the manuscript and appendix, including fixing grammatical issues, improving the clarity and fluency of non-native expressions.
We did not use LLMs to generate research ideas, design methods, conduct experiments, search for related work, or produce scientific claims.
All technical contributions, experimental results, analyses, and implementations are original and have been verified by the authors.

\subsection{Reproducibility}

We will publicly release our code in
\url{https://github.com/yikai-zhu/ConRAG}
to help reproduce the experimental results of this paper.

%% file: Tables/dataset_statistics.tex
\begin{table}[t]
    \centering
    \setlength{\tabcolsep}{11pt}
    \resizebox{0.45\textwidth}{!}{%
    \begin{tabular}{@{}lcc@{}}
        \toprule
        \bf Dataset & \bf \#Passages &\bf \#Question Types \\
        \midrule
        HotpotQA        & 9,811  & 2 \\
        2WikiMultiHopQA & 6,119  & 4 \\
        MuSiQue         & 11,656 & 3 \\
        \bottomrule
    \end{tabular}%
    }
    \caption{
    \textbf{Statistics of all evaluated datasets.} ``\#Passages'' denotes the number of passages in the retrieval corpus, and ``\#Question Types'' denotes the number of question categories covered by each benchmark.}
    \label{tab:dataset_statistics}
\end{table}

%% file: Tables/consensus_ablation_analysis.tex
\begin{table*}[t]
\centering
\setlength{\tabcolsep}{12pt}
\resizebox{\textwidth}{!}{%
\begin{tabular}{@{}lcccccccc@{}}
\toprule
\multirow{2}{*}{\textbf{Variant}}
& \multicolumn{2}{c}{\textbf{HotpotQA}}
& \multicolumn{2}{c}{\textbf{2WikiMultiHopQA}}
& \multicolumn{2}{c}{\textbf{MuSiQue}}
& \multicolumn{2}{c}{\textbf{Overall}} \\
\cmidrule(lr){2-3}
\cmidrule(lr){4-5}
\cmidrule(lr){6-7}
\cmidrule(lr){8-9}
& \textbf{Str-Acc} & \textbf{LLM-Acc}
& \textbf{Str-Acc} & \textbf{LLM-Acc}
& \textbf{Str-Acc} & \textbf{LLM-Acc}
& \textbf{Avg.} & $\boldsymbol{\Delta}\uparrow$ \\
\midrule
\textbf{Full \method{}}
& \textbf{61.3} & \textbf{68.3}
& \textbf{74.2} & \textbf{74.1}
& \textbf{38.9} & \textbf{43.8}
& \textbf{60.1} & -- \\
\hdashline
\quad -w/o Entity Specificity
& 60.1 & 67.4
& 74.1 & 73.9
& 38.7 & 43.3
& 59.6 & $-0.5$ \\
\quad -w/o Relation Residual Pooling
& 60.3 & 67.5
& 73.2 & 71.9
& 38.7 & 43.7
& 59.2 & $-0.9$ \\
\bottomrule
\end{tabular}%
}
\caption{\textbf{Ablation study of scoring mechanisms in \textit{Consensus}} using \texttt{GPT-4o-mini}.
``Avg.'' denotes the average results and ``$\Delta\uparrow$'' refers to the changes relative to Full \method{}.}

\label{tab:consensus_ablation}
\end{table*}

%% file: Tables/kc_sensitivity.tex
\begin{table*}[t]
\centering
\setlength{\tabcolsep}{15pt}
\resizebox{\textwidth}{!}{%
\begin{tabular}{@{}lcccccccc@{}}
\toprule
\multirow{2}{*}{$\boldsymbol{K_c}$}
& \multicolumn{2}{c}{\textbf{HotpotQA}}
& \multicolumn{2}{c}{\textbf{2WikiMultiHopQA}}
& \multicolumn{2}{c}{\textbf{MuSiQue}}
& \multicolumn{2}{c}{\textbf{Average Score}} \\
\cmidrule(lr){2-3}
\cmidrule(lr){4-5}
\cmidrule(lr){6-7}
\cmidrule(lr){8-9}
& \textbf{Str-Acc} & \textbf{LLM-Acc}
& \textbf{Str-Acc} & \textbf{LLM-Acc}
& \textbf{Str-Acc} & \textbf{LLM-Acc}
& \textbf{Str-Acc} & \textbf{LLM-Acc} \\
\midrule
8  & 59.9 & 66.8 & 73.0 & 73.0 & 39.9 & 45.2 & 57.6 & 61.7 \\
16 & 61.3 & 68.3 & 74.2 & 74.1 & 38.9 & 43.8 & 58.1 & 62.1 \\
24 & 60.4 & 68.2 & 74.2 & 74.0 & 39.5 & 44.2 & 58.0 & 62.1 \\
32 & 60.7 & 67.0 & 74.0 & 73.7 & 38.8 & 43.7 & 57.8 & 61.5 \\
\bottomrule
\end{tabular}%
}
\caption{\textbf{Parameter analysis of the candidate budget $K_c$} using \texttt{GPT-4o-mini}.
$K_c$ limits the number of retrieved objects per view, with the final passage budget fixed at $K_f=3$ per retrieval.
}
\label{tab:kc_sensitivity}
\end{table*}

%% file: Tables/additional_metrics.tex
\begin{table*}[t]
    \centering
    \setlength{\tabcolsep}{14pt}
    \resizebox{\textwidth}{!}{%
    \begin{tabular}{llcccccccc}
        \toprule
        \multirow{2}{*}{\textbf{Type}}
        & \multirow{2}{*}{\textbf{Method}}
        & \multicolumn{2}{c}{\textbf{HotpotQA}}
        & \multicolumn{2}{c}{\textbf{2WikiMultiHopQA}}
        & \multicolumn{2}{c}{\textbf{MuSiQue}}
        & \multicolumn{2}{c}{\textbf{Overall}} \\
        \cmidrule(lr){3-4}
        \cmidrule(lr){5-6}
        \cmidrule(lr){7-8}
        \cmidrule(lr){9-10}
        &
        & \textbf{EM} & \textbf{F1}
        & \textbf{EM} & \textbf{F1}
        & \textbf{EM} & \textbf{F1}
        & \textbf{Avg.} & $\boldsymbol{\Delta}\uparrow$ \\
        \midrule

        \multirow{2}{*}{Standard} 
         & Zero-shot & 26.4 & 35.7 & 31.0 & 35.8 & 6.0 & 15.1 & 25.0 & $-10.5$ \\
         & Vanilla RAG & 41.3 & 51.6 & 36.1 & 39.6 & 17.9 & 26.4 & 35.5 & -- \\
        \midrule

        \multirow{4}{*}{Corpus-side}
         & LightRAG & 45.6 & 55.7 & 28.2 & 34.6 & 19.5 & 28.1 & 35.3 & $-0.2\phantom{0}$ \\
         & HippoRAG & 48.6 & 59.2 & 52.2 & 58.8 & 21.3 & 29.8 & 45.0 & $+9.5\phantom{0}$ \\
         & HippoRAG~2 & 47.7 & 58.9 & 44.7 & 51.7 & 21.5 & 29.3 & 42.3 & $+6.8\phantom{0}$ \\
         & LinearRAG & 40.6 & 49.5 & 41.9 & 47.3 & 14.7 & 21.6 & 35.9 & $+0.4\phantom{0}$ \\
        \midrule

        \multirow{1}{*}{Query-side}
         & LogicRAG & 54.2 & 66.3 & 59.7 & 67.8 & 26.7 & 37.8 & 52.1 & $+16.6$ \\
        \midrule

        \multirow{2}{*}{Hybrid}
         & ChainRAG & 52.5 & 64.6 & 46.4 & 53.5 & 34.8 & 45.4 & 49.5 & $+14.0$ \\
         & Youtu-GraphRAG & 60.8 & 73.7 & 68.7 & 76.5 & 39.5 & 51.5 & 61.8 & $+26.3$ \\
        \midrule

        Ours & \textbf{\method{}} & \textbf{61.5} & \textbf{74.3} & \textbf{69.9} & \textbf{78.2} & \textbf{40.4} & \textbf{53.1} & \textbf{62.9} & \makebox[0pt][l]{$\mathbf{+27.4}$}\phantom{$+27.4$} \\
        \bottomrule
    \end{tabular}%
    }
    \caption{\textbf{Performance comparison (\%) on EM and F1 across three benchmarks} using \texttt{Gemma-4-31B-it}.
    ``Avg.'' denotes the average results and ``$\Delta\uparrow$'' refers to the gains against Vanilla RAG. Best results are in \textbf{bold}.}
    \label{tab:additional_metrics}
\end{table*}

%% file: Tables/stage_wise_efficiency.tex
\begin{table*}[t]
    \centering
    \setlength{\tabcolsep}{12pt}
    \resizebox{\textwidth}{!}{%
        \begin{tabular}{@{}llcccccc@{}}
            \toprule
            \multirow{2}{*}{\textbf{Type}}
            & \multirow{2}{*}{\textbf{Method}}
            & \multicolumn{2}{c}{\textbf{Offline Construction}}
            & \multicolumn{2}{c}{\textbf{Online Inference}}
            & \multicolumn{2}{c}{\textbf{Total Cost}} \\
            \cmidrule(lr){3-4}\cmidrule(lr){5-6}\cmidrule(lr){7-8}
            &
            & \textbf{Time/q (s)} & \textbf{Tokens/q}
            & \textbf{Time/q (s)} & \textbf{Tokens/q}
            & \textbf{Time/q (s)} & \textbf{Tokens/q} \\
            \midrule

            \multirow{2}{*}{Standard}
            & Zero-shot
            & 0.0 & 0.0 & 0.1 & 158.4 & 0.1 & 158.4 \\
            & Vanilla RAG
            & 0.0 & 0.0 & 0.2 & 591.3 & 0.2 & 591.3 \\
            \midrule

            \multirow{4}{*}{Corpus-side}
            & LightRAG
            & 4.7 & 40,025.4 & 1.6 & 3,199.3 & 6.3 & 43,224.7 \\
            & HippoRAG
            & 4.0 & 8,679.6 & 4.5 & 2,214.3 & 8.5 & 10,893.9 \\
            & HippoRAG~2
            & 8.1 & 8,067.4 & 2.6 & 3,799.7 & 10.7 & 11,867.1 \\
            & LinearRAG
            & 0.4 & 0.0 & 0.2 & 669.9 & 0.6 & 669.9 \\
            \midrule

            Query-side
            & LogicRAG
            & 0.0 & 0.0 & 18.7 & 3,049.4 & 18.7 & 3,049.4 \\
            \midrule

            \multirow{2}{*}{Hybrid}
            & ChainRAG
            & 0.3 & 0.0 & 1.7 & 3,157.1 & 2.0 & 3,157.1 \\
            & Youtu-GraphRAG
            & 1.4 & 14,949.1 & 14.1 & 9,665.8 & 15.5 & 24,614.9 \\
            \midrule

            Ours
            & \textbf{ConRAG}
            & 1.6 & 11,951.7 & 2.7 & 2,362.1 & 4.3 & 14,313.8 \\
            \bottomrule
        \end{tabular}%
    }
    \caption{\textbf{Stage-wise efficiency comparison of RAG methods on 2WikiMultiHopQA} using \texttt{Gemma-4-31B-it}. We report time and token costs per question separately for offline construction and online inference, with offline costs amortized over all evaluation questions.}
    \label{tab:stage_wise_efficiency}
\end{table*}

%% file: Tables/component_ablation.tex
\begin{table*}[t]
    \centering
    \setlength{\tabcolsep}{14pt}
    \resizebox{1\textwidth}{!}{%
    \begin{tabular}{@{}lcccccccc@{}}
        \toprule
        \multirow{2}{*}{\textbf{Variant}}
        & \multicolumn{2}{c}{\textbf{HotpotQA}}
        & \multicolumn{2}{c}{\textbf{2WikiMultiHopQA}}
        & \multicolumn{2}{c}{\textbf{MuSiQue}}
        & \multicolumn{2}{c}{\textbf{Overall}} \\
        \cmidrule(lr){2-3}
        \cmidrule(lr){4-5}
        \cmidrule(lr){6-7}
        \cmidrule(lr){8-9}
        & \textbf{Str-Acc} & \textbf{LLM-Acc}
        & \textbf{Str-Acc} & \textbf{LLM-Acc}
        & \textbf{Str-Acc} & \textbf{LLM-Acc}
        & \textbf{Avg.} & $\boldsymbol{\Delta}\uparrow$ \\
        \midrule
        \textbf{Full \method{}}
        & \textbf{61.3} & \textbf{68.3}
        & \textbf{74.2} & \textbf{74.1}
        & \textbf{38.9} & \textbf{43.8}
        & \textbf{60.1} & -- \\
        \hdashline
        \quad -w/o \textit{Consensus}
        & 59.1 & 65.4
        & 72.1 & 70.7
        & 35.5 & 40.3
        & 57.2 & $-2.9$ \\
        \quad -w/o \textit{Constraint}
        & 59.2 & 66.0
        & 72.7 & 71.9
        & 37.1 & 42.6
        & 58.3 & $-1.8$ \\
        \bottomrule
    \end{tabular}%
    }
    \caption{\textbf{Full results of Figure~\ref{fig:component_ablation}, \textit{i.e.}, ablation study of \textit{Consensus} and \textit{Constraint}.}
    ``Avg.'' denotes the average results and ``$\Delta\uparrow$'' refers to the changes relative to Full \method{}.}
    \label{tab:component_ablation}
\end{table*}

%% file: Tables/view_ablation_full.tex
\begin{table*}[ht]
    \centering
    \setlength{\tabcolsep}{12pt}
    \resizebox{1\textwidth}{!}{%
    \begin{tabular}{@{}w{c}{1cm} w{c}{1cm} w{c}{1cm}cccccccc@{}}
        \toprule
        \multicolumn{3}{c}{\textbf{Retrieval Views}}
        & \multicolumn{2}{c}{\textbf{HotpotQA}}
        & \multicolumn{2}{c}{\textbf{2WikiMultiHopQA}}
        & \multicolumn{2}{c}{\textbf{MuSiQue}}
        & \multicolumn{2}{c}{\textbf{Average Score}} \\
        \cmidrule(lr){1-3}
        \cmidrule(lr){4-5}
        \cmidrule(lr){6-7}
        \cmidrule(lr){8-9}
        \cmidrule(lr){10-11}
        \textbf{Text} & \textbf{Entity} & \textbf{Relation}
        & \textbf{Str-Acc} & \textbf{LLM-Acc}
        & \textbf{Str-Acc} & \textbf{LLM-Acc}
        & \textbf{Str-Acc} & \textbf{LLM-Acc}
        & \textbf{Str-Acc} & \textbf{LLM-Acc} \\
        \midrule

        \cmark & \xmark & \xmark
        & 59.3 & 65.8
        & 71.2 & 70.0
        & 37.7 & 42.1
        & 56.1 & 59.3 \\

        \xmark & \cmark & \xmark
        & 47.1 & 52.0
        & 59.3 & 59.0
        & 27.3 & 31.7
        & 44.6 & 47.6 \\

        \xmark & \xmark & \cmark
        & 53.6 & 59.7
        & 71.0 & 69.6
        & 31.5 & 36.5
        & 52.0 & 55.3 \\

        \cmark & \cmark & \xmark
        & 59.5 & 66.7
        & 71.7 & 70.6
        & 38.6 & 43.3
        & 56.6 & 60.2 \\

        \cmark & \xmark & \cmark
        & 59.6 & 66.9
        & 74.0 & 73.2
        & 38.7 & 43.5
        & 57.4 & 61.2 \\

        \xmark & \cmark & \cmark
        & 53.8 & 60.7
        & 69.7 & 68.1
        & 30.4 & 35.3
        & 51.3 & 54.7 \\

        \cmark & \cmark & \cmark
        & \textbf{61.3} & \textbf{68.3}
        & \textbf{74.2} & \textbf{74.1}
        & \textbf{38.9} & \textbf{43.8}
        & \textbf{58.1} & \textbf{62.1} \\
        \bottomrule
    \end{tabular}%
    }
    \caption{\textbf{Full results of Table~\ref{tab:view_ablation}, \textit{i.e.}, ablation study on different retrieval-view combinations in \method{}.}
    Unlike Table~\ref{tab:view_ablation}, which reports averages across benchmarks, this table reports results for each benchmark.
    }
    
    \label{tab:view_ablation_full}
\end{table*}

%% file: Tables/full_fusion_comparison.tex
\begin{table*}[t]
\centering
\setlength{\tabcolsep}{16pt}
\resizebox{\textwidth}{!}{%
\begin{tabular}{@{}lcccccccc@{}}
\toprule
\multirow{2}{*}{\textbf{Fusion}}
& \multicolumn{2}{c}{\textbf{HotpotQA}}
& \multicolumn{2}{c}{\textbf{2WikiMultiHopQA}}
& \multicolumn{2}{c}{\textbf{MuSiQue}}
& \multicolumn{2}{c}{\textbf{Overall}} \\
\cmidrule(lr){2-3}
\cmidrule(lr){4-5}
\cmidrule(lr){6-7}
\cmidrule(lr){8-9}
& \textbf{Str-Acc} & \textbf{LLM-Acc}
& \textbf{Str-Acc} & \textbf{LLM-Acc}
& \textbf{Str-Acc} & \textbf{LLM-Acc}
& \textbf{Avg.} & $\boldsymbol{\Delta}\uparrow$ \\
\midrule
Equal-mean & 59.1 & 65.4 & 72.1 & 70.7 & 35.5 & 40.3 & 57.2 & $-2.9$ \\
CombMNZ    & 57.9 & 64.9 & 72.5 & 71.2 & 34.8 & 40.5 & 57.0 & $-3.1$ \\
RRF        & 57.4 & 64.9 & 71.1 & 69.7 & 33.0 & 38.0 & 55.7 & $-4.4$ \\
RRA        & 57.8 & 64.6 & 71.5 & 70.5 & 36.8 & 41.3 & 57.1 & $-3.0$ \\
\midrule
\textbf{\textit{Consensus}}
& \textbf{61.3} & \textbf{68.3}
& \textbf{74.2} & \textbf{74.1}
& \textbf{38.9} & \textbf{43.8}
& \textbf{60.1} & -- \\
\bottomrule
\end{tabular}%
}
\caption{\textbf{Full results of Table~\ref{tab:fusion_strategies_main}, \textit{i.e.}, performance comparison (\%) of fusion strategies in \textit{Consensus}} using \texttt{GPT-4o-mini}.
``Avg.'' denotes the average results and ``$\Delta\uparrow$'' refers to the changes relative to Full \method{}.}
\label{tab:fusion_comparison}
\end{table*}

%% file: Tables/slot_binding_analysis.tex
\begin{table*}[t]
\centering
\resizebox{\textwidth}{!}{%
\begin{tabular}{@{}lcccccccccc@{}}
\toprule
\multirow{2}{*}{\textbf{Hop count}}
& \multirow{2}{*}{\textbf{\#Questions}}
& \multicolumn{2}{c}{\textbf{Binding success}}
& \multicolumn{2}{c}{\textbf{No slots}}
& \multicolumn{2}{c}{\textbf{Binding failure}}
& \multicolumn{3}{c}{\textbf{All questions}} \\
\cmidrule(lr){3-4}
\cmidrule(lr){5-6}
\cmidrule(lr){7-8}
\cmidrule(lr){9-11}
& & \textbf{\#Questions} & \textbf{LLM-Acc}
& \textbf{\#Questions} & \textbf{LLM-Acc}
& \textbf{\#Questions} & \textbf{LLM-Acc}
& \textbf{Full \method{}} & \textbf{w/o \textit{Constraint}}
& $\boldsymbol{\Delta \uparrow}$ \\
\midrule
2-hop & 518 & 93 & 48.4 & 424 & 59.9 & 1 & 0.0
& 57.7 & 55.8 & $+1.9\phantom{0}$ \\
3-hop & 316 & 196 & 58.7 & 118 & 50.0 & 2 & 100.0
& 55.7 & 43.0 & $+12.7$ \\
4-hop & 166 & 146 & 36.3 & 18 & 22.2 & 2 & 50.0
& 34.9 & 21.7 & $+13.2$ \\
\midrule
Overall & 1,000
& 435 & 49.0
& 560 & 56.6
& 5 & 60.0
& 53.3 & 46.1 & $+7.2\phantom{0}$ \\
\bottomrule
\end{tabular}%
}
\caption{\textbf{Full results of Figure~\ref{fig:slot_binding_analysis}, \textit{i.e.}, the effect of slot binding across reasoning depths on MuSiQue.}
``$\Delta\uparrow$'' refers to the LLM-Acc gains of Full \method{} relative to ``w/o \textit{Constraint}''.}
\label{tab:slot_binding_analysis}
\end{table*}

%% file: Tables/question_type_analysis.tex
\begin{table*}[t]
\centering
\setlength{\tabcolsep}{20pt}
\resizebox{1\textwidth}{!}{%
\begin{tabular}{@{}llccccc@{}}
\toprule
\multirow{2}{*}{\textbf{Dataset}}
& \multirow{2}{*}{\textbf{Type}}
& \multirow{2}{*}{\textbf{\#Questions}}

& \multicolumn{2}{c}{\textbf{\method{}}}
& \multicolumn{2}{c}{\textbf{w/o \textit{Consensus}}} \\
\cmidrule(lr){4-5}
\cmidrule(lr){6-7}
& & & \textbf{Str-Acc} & \textbf{LLM-Acc}
& \textbf{Str-Acc} & \textbf{LLM-Acc} \\
\midrule
\multirow{3}{*}{\textbf{HotpotQA}}
& Bridge
& 811 & 57.2 & 64.7 & 56.0 & 62.8 \\
& Comparison
& 189 & 78.8 & 83.6 & 72.5 & 76.7 \\
& \textbf{Overall}
& 1,000 & 61.3 & 68.3 & 59.1 & 65.4 \\
\midrule
\multirow{5}{*}{\textbf{2WikiMultiHopQA}}
& Bridge-comparison
& 235 & 75.3 & 76.2 & 67.2 & 67.2 \\
& Comparison
& 244 & 90.2 & 91.4 & 88.9 & 88.9 \\
& Compositional
& 413 & 67.1 & 63.9 & 68.3 & 63.4 \\
& Inference
& 108 & 63.0 & 69.4 & 59.3 & 64.8 \\
& \textbf{Overall}
& 1,000 & 74.2 & 74.1 & 72.1 & 70.7 \\
\midrule
\multirow{4}{*}{\textbf{MuSiQue}}
& 2-hop
& 518 & 49.2 & 54.1 & 46.5 & 51.4 \\
& 3-hop
& 316 & 32.9 & 38.6 & 29.7 & 35.1 \\
& 4-hop
& 166 & 18.1 & 21.7 & 12.0 & 15.7 \\
& \textbf{Overall}
& 1,000 & 38.9 & 43.8 & 35.5 & 40.3 \\
\bottomrule
\end{tabular}%
}
\caption{\textbf{Full results of Figure~\ref{fig:question_type_analysis}, \textit{i.e.}, performance by question type and reasoning depth.}
\#Questions is the number of questions in each group.}
\label{tab:question_type_analysis}
\vspace{20pt}
\end{table*}

%% file: Tables/top-k_analysis.tex
\begin{table*}[t]
\centering
\setlength{\tabcolsep}{10pt}
\resizebox{\textwidth}{!}{%
\begin{tabular}{@{}lccccccccccc@{}}
\toprule
\multirow{2}{*}{$\boldsymbol{K_f}$}
& \multicolumn{3}{c}{\textbf{HotpotQA}}
& \multicolumn{3}{c}{\textbf{2WikiMultiHopQA}}
& \multicolumn{3}{c}{\textbf{MuSiQue}}
& \multicolumn{2}{c}{\textbf{Overall}} \\
\cmidrule(lr){2-4}
\cmidrule(lr){5-7}
\cmidrule(lr){8-10}
\cmidrule(lr){11-12}
& \textbf{Str-Acc}
& \textbf{LLM-Acc}
& \textbf{Tokens/q}
& \textbf{Str-Acc}
& \textbf{LLM-Acc}
& \textbf{Tokens/q}
& \textbf{Str-Acc}
& \textbf{LLM-Acc}
& \textbf{Tokens/q}
& \textbf{Accuracy}
& \textbf{Tokens/q} \\
\midrule

1  & 54.3 & 61.2 & 1,274.0 & 69.3 & 67.7 & 1,429.0 & 32.3 & 35.6 & 1,465.8 & 53.4 & 1,389.6 \\
3  & 61.3 & 68.3 & 1,846.0 & 74.2 & 74.1 & 2,110.6 & 38.9 & 43.8 & 2,028.6 & 60.1 & 1,995.1 \\
5  & 61.9 & 68.9 & 2,398.6 & 75.0 & 74.6 & 2,703.2 & 41.3 & 45.6 & 2,649.8 & 61.2 & 2,583.9 \\
8  & 61.9 & 69.5 & 3,295.5 & 74.9 & 74.4 & 3,649.5 & 41.7 & 48.0 & 3,476.9 & 61.7 & 3,474.0 \\
10 & 62.4 & 70.3 & 3,901.7 & 75.9 & 74.9 & 4,354.2 & 42.4 & 48.7 & 4,077.7 & 62.4 & 4,111.2 \\
\bottomrule
\end{tabular}%
}
\caption{\textbf{Full results of Figure~\ref{fig:kf_tradeoff}, \textit{i.e.}, quality-efficiency trade-off with different passage budgets $K_f$} using \texttt{GPT-4o-mini}.
$K_f$ is the passage budget per retrieval.
``Tokens/q'' denotes online token consumption per question.}
\label{tab:top-k_analysis}
\end{table*}

%% file: Tables/efficiency_analysis.tex
\begin{table*}[t]
    \centering
    \resizebox{\textwidth}{!}{%
        \begin{tabular}{@{}llcccccccc@{}}
            \toprule
            \multirow{2}{*}{\textbf{Type}}
            & \multirow{2}{*}{\textbf{Method}}
            & \multicolumn{2}{c}{\textbf{HotpotQA}}
            & \multicolumn{2}{c}{\textbf{2WikiMultiHopQA}}
            & \multicolumn{2}{c}{\textbf{MuSiQue}}
            & \multicolumn{2}{c}{\textbf{Average Cost}} \\
            \cmidrule(lr){3-4}\cmidrule(lr){5-6}\cmidrule(lr){7-8}\cmidrule(lr){9-10}
            &
            & \textbf{Time/q (s)} & \textbf{Tokens/q}
            & \textbf{Time/q (s)} & \textbf{Tokens/q}
            & \textbf{Time/q (s)} & \textbf{Tokens/q}
            & \textbf{Time/q (s)} & \textbf{Tokens/q} \\
            \midrule

            \multirow{2}{*}{Standard}
            & Zero-shot
            & 0.2 & 162.0 & 0.1 & 158.4 & 0.1 & 164.7 & 0.1 & 161.7 \\
            & Vanilla RAG
            & 0.2 & 586.1 & 0.2 & 591.3 & 0.3 & 619.9 & 0.2 & 599.1 \\
            \midrule

            \multirow{4}{*}{Corpus-side}
            & LightRAG
            & 9.5 & 71,990.1 & 6.3 & 43,224.7 & 11.3 & 85,445.2 & 9.0 & 66,886.7 \\
            & HippoRAG
            & 9.1 & 17,662.7 & 8.5 & 10,893.9 & 9.5 & 20,269.5 & 9.0 & 16,275.4 \\
            & HippoRAG~2
            & 11.4 & 17,540.2 & 10.7 & 11,867.1 & 12.5 & 19,718.2 & 11.5 & 16,375.2 \\
            & LinearRAG
            & 0.5 & 642.8 & 0.6 & 669.9 & 1.0 & 719.9 & 0.7 & 677.5 \\
            \midrule

            \multirow{1}{*}{Query-side}
            & LogicRAG
            & 19.5 & 2,443.0 & 18.7 & 3,049.4 & 30.1 & 3,695.1 & 22.8 & 3,062.5 \\
            \midrule

            \multirow{2}{*}{Hybrid}
            & ChainRAG
            & 2.1 & 3,012.3
            & 2.0 & 3,157.1
            & 2.7 & 3,637.1
            & 2.3 & 3,268.8 \\
            & Youtu-GraphRAG
            & 20.0 & 52,428.6
            & 15.5 & 24,614.9
            & 36.8 & 111,008.2
            & 24.1 & 62,683.9 \\
            \midrule

            Ours
            & \textbf{\method{}}
            & 5.2 & 21,737.6 & 4.3 & 14,313.8 & 7.8 & 25,422.4 & 5.8 & 20,491.3 \\
            \bottomrule
        \end{tabular}%
    }
    \caption{\textbf{Efficiency comparison of different RAG methods across three benchmarks} using \texttt{Gemma-4-31B-it}.
    Time and token costs cover the full pipeline and are averaged per question.}
    \label{tab:efficiency_analysis}
\end{table*}

%% file: Tables/case_study.tex
\begin{table*}[t]
\centering
\small
\setlength{\tabcolsep}{5pt}
\renewcommand{\arraystretch}{1.12}
\begin{tabular}{@{}p{\textwidth}@{}}
\toprule
\textbf{Question:} Who won the indy car race in the largest populated city in the state that the performer of Changes Two is from? \\
\textbf{Gold answer:} \textbf{Mario Andretti} \\
\textbf{Evidence chain:} \textit{Changes Two} $\rightarrow$ \textbf{Charles Mingus} $\rightarrow$ \textbf{Arizona} $\rightarrow$ \textbf{Phoenix} $\rightarrow$ \textbf{Mario Andretti} \\
\bottomrule
\end{tabular}

\smallskip
\begin{tabular}{@{}lp{0.21\textwidth}p{0.72\textwidth}@{}}
\toprule
\multicolumn{3}{@{}l}{\textbf{(a) Supporting passages}} \\
\midrule
\textbf{P1} & \textit{Changes Two} & Changes Two is an album by \textbf{Charles Mingus}. \\
\textbf{P2} & \textit{Charles Mingus} & Charles Mingus was born in Nogales, \textbf{Arizona}. \\
\textbf{P3} & \textit{Tucson, Arizona} & Tucson is the second-largest populated city in \textbf{Arizona} behind \textbf{Phoenix}, both of which anchor the Arizona Sun Corridor. \\
\textbf{P4} & \textit{Desert Diamond West Valley Phoenix Grand Prix} & \textbf{Phoenix} has a rich history of open wheel races, including [\ldots] the final career victory for Indy legend \textbf{Mario Andretti} (1993). \\
\bottomrule
\end{tabular}

\smallskip
\begin{tabular*}{\textwidth}{@{}p{0.30\textwidth}@{\extracolsep{\fill}}cccccccc@{}}
\toprule
\multicolumn{9}{@{}l}{\textbf{(b) \textit{Consensus}: original-question passage ranking}} \\
\midrule
\multirow{2}{*}{\textbf{Passage}} & \multicolumn{5}{c}{\textbf{Score ($\times100$)}} & \multicolumn{3}{c}{\textbf{Rank}} \\
\cmidrule(lr){2-6}\cmidrule(lr){7-9}
& $T$ & $E$ & $R$ & \textbf{Mean} & $S$ & \textbf{Text} & \textbf{Mean} & \textbf{\textit{Consensus}} \\
\midrule
2018 Indianapolis 500 & 100.0 & 71.5 & 94.0 & 88.5 & 145.3 & \textbf{1} & \textbf{1} & \textbf{1} \\
Indiana & 56.9 & 2.0 & 17.5 & 25.5 & 65.8 & \textbf{2} & 5 & \textbf{2} \\
\textbf{P4: Desert Diamond West Valley Phoenix Grand Prix} & \textbf{43.9} & \textbf{30.8} & \textbf{9.3} & \textbf{28.0} & \textbf{60.6} & 4 & 4 & \textbf{3} \\
Elmer George & 16.2 & 100.0 & 24.9 & 47.0 & 54.6 & 12 & \textbf{2} & 4 \\
List of Daytona 500 pole position winners & 47.1 & 0.0 & 0.0 & 15.7 & 47.1 & \textbf{3} & 8 & 5 \\
Richmond, Virginia & 11.2 & 3.8 & 100.0 & 38.3 & 45.4 & 13 & \textbf{3} & 6 \\
\bottomrule
\end{tabular*}
\caption{\textbf{Case study of \textit{Consensus}.}
$T$, $E$, and $R$ denote Text, Entity, and Relation scores; Mean and $S$ denote equal-mean and \textit{Consensus} scores.
Bold ranks indicate top-3 selection.}
\label{tab:case_study}
\end{table*}

\begin{table*}[t]
\centering
\small
\setlength{\tabcolsep}{3pt}
\renewcommand{\arraystretch}{1.02}
\begin{tabular}{@{}p{0.16\textwidth}>{\raggedright\arraybackslash}p{0.40\textwidth}>{\raggedright\arraybackslash}p{0.40\textwidth}@{}}
\toprule
\textbf{Stage} & \textbf{w/o \textit{Constraint}} & \textbf{Full \method{}} \\
\midrule
\multicolumn{3}{@{}l}{\textbf{Shared initial retrieval and step 0}, $D_0=[\,]$} \\
Initial top-3 & \multicolumn{2}{p{0.80\textwidth}@{}}{2018 Indianapolis 500; Indiana; \textbf{P4}} \\
Initial summary $I_0$ & \multicolumn{2}{p{0.80\textwidth}@{}}{The largest city in Indiana is Indianapolis. Will Power won the 2018 Indianapolis 500} \\
$q_0$ / retrieval query & \multicolumn{2}{p{0.80\textwidth}@{}}{Who is the performer of the work `Changes Two'?} \\
Step 0 top-3 & \multicolumn{2}{p{0.80\textwidth}@{}}{\textbf{P1}; 2 Horns / 2 Rhythm; Both Sides, Now} \\
Answer $a_0$ & \multicolumn{2}{p{0.80\textwidth}@{}}{\textbf{Charles Mingus}} \\
Summary $m_0$ & \multicolumn{2}{p{0.80\textwidth}@{}}{The performer of the album `Changes Two' is \textbf{Charles Mingus}. The largest city in Indiana is Indianapolis. Will Power won the 2018 Indianapolis 500} \\
\midrule
\multicolumn{3}{@{}l}{\textbf{Step 1: performer $\rightarrow$ state}, $D_1=[0]$} \\
$q_1$ & Which state is the performer of `Changes Two' from? & Which state is the performer \texttt{<dep:0>} from? \\
Retrieval query & Same as $q_1$ & Which state is the performer \textbf{Charles Mingus} from? \\
Slot binding & Not applicable & \texttt{<dep:0>} $\mapsto$ \textbf{Charles Mingus} (success) \\
Top-3 passages & \textbf{P1}; Article Five of the United States Constitution; Election 2 & Mingus Plays Piano; \textbf{P2}; Mingus Three \\
Answer $a_1$ & I am sorry, but the provided evidence does not state which state \textbf{Charles Mingus} is from & \textbf{Arizona} \\
Summary $m_1$ & The performer of the album `Changes Two' is \textbf{Charles Mingus} & The performer of the album `Changes Two' is \textbf{Charles Mingus}, who was born in Nogales, \textbf{Arizona} \\
\midrule
\multicolumn{3}{@{}l}{\textbf{Step 2: state $\rightarrow$ city}, $D_2=[1]$} \\
$q_2$ & What is the largest populated city in the state the performer of `Changes Two' is from? & What is the largest populated city in the state \texttt{<dep:1>}? \\
Retrieval query & Same as $q_2$ & What is the largest populated city in the state \textbf{Arizona}? \\
Slot binding & Not applicable & \texttt{<dep:1>} $\mapsto$ \textbf{Arizona} (success) \\
Top-3 passages & \textbf{P1}; Oklahoma; Black people & Arizona; Tucson, Arizona (other passage); \textbf{P3} \\
Answer $a_2$ & Oklahoma City & \textbf{Phoenix} \\
Summary $m_2$ & The performer of `Changes Two' is \textbf{Charles Mingus}. \textbf{Charles Mingus} is from Oklahoma. The largest populated city in Oklahoma is Oklahoma City & The performer of the album `Changes Two' is \textbf{Charles Mingus}, who was born in Nogales, \textbf{Arizona}. The largest populated city in \textbf{Arizona} is \textbf{Phoenix} \\
\midrule
\multicolumn{3}{@{}l}{\textbf{Step 3: city $\rightarrow$ race winner}, $D_3=[2]$} \\
$q_3$ & Who won the Indy car race in the largest populated city in the state the performer of `Changes Two' is from? & Who won the Indy car race in the city \texttt{<dep:2>}? \\
Retrieval query & Same as $q_3$ & Who won the Indy car race in the city \textbf{Phoenix}? \\
Slot binding & Not applicable & \texttt{<dep:2>} $\mapsto$ \textbf{Phoenix} (success) \\
Top-3 passages & 2018 Indianapolis 500; \textbf{P4}; Indiana & \textbf{P4}; 2018 Indianapolis 500; Elmer George \\
Answer $a_3$ & I do not have information on who won an Indy car race in Oklahoma City & \textbf{Mario Andretti} \\
Summary $m_3$ & The performer of `Changes Two' is \textbf{Charles Mingus}, who is from Oklahoma. The largest populated city in Oklahoma is Oklahoma City & The performer of `Changes Two' is \textbf{Charles Mingus}, who is from Nogales, \textbf{Arizona}. The largest populated city in \textbf{Arizona} is \textbf{Phoenix}. \textbf{Mario Andretti} won the final career victory for Indy legend in \textbf{Phoenix} in 1993 \\
\midrule
Final answer & I do not have information on who won an Indy car race in Oklahoma City & \textbf{Mario Andretti} \\
\bottomrule
\end{tabular}
\caption{\textbf{Case study of \textit{Constraint}.}
We compare Full \method{} with the ``w/o \textit{Constraint}'' variant on the MuSiQue question in Table~\ref{tab:case_study} using \texttt{Gemma-4-31B-it}.
$D_i$, $a_i$, and $m_i$ denote the dependencies, answer, and evidence summary of step $i$, respectively.
Step memory contains $I_0$ and nonempty evidence summaries from direct predecessors.}
\label{tab:case_study_constraint}
\end{table*}

%% file: Figures/prompt_figures.tex
\lstdefinestyle{conragprompt}{
    basicstyle=\scriptsize\ttfamily,
    columns=fullflexible,
    keepspaces=true,
    breaklines=true,
    breakatwhitespace=true,
    showstringspaces=false,
    upquote=true,
    numbers=none,
    frame=none,
    xleftmargin=0pt,
    xrightmargin=0pt,
    aboveskip=0pt,
    belowskip=0pt
}
\definecolor{ConRAGPromptBlue}{HTML}{4273B1}
\definecolor{ConRAGPromptBackground}{HTML}{F8F8F8}
\tcbset{conragpromptbox/.style={
    enhanced,
    listing only,
    listing options={style=conragprompt},
    colback=ConRAGPromptBackground,
    colframe=ConRAGPromptBlue,
    colbacktitle=ConRAGPromptBlue,
    coltitle=white,
    fonttitle=\bfseries\small,
    boxrule=0.6pt,
    arc=2mm,
    left=2pt,
    right=2pt,
    top=2pt,
    bottom=2pt,
    toptitle=1pt,
    bottomtitle=1pt,
    before skip=0pt,
    after skip=0pt
}}

\begin{figure*}[ht]
    \centering
    \begin{minipage}[t]{\textwidth}
    \input{Figures/prompt_knowledge_graph_extraction}
    \end{minipage}
\end{figure*}

\begin{figure*}[ht]
    \centering
    \begin{minipage}[t]{\textwidth}
    \input{Figures/knowledge_graph_schema}
    \end{minipage}
\end{figure*}

\begin{figure*}[ht]
    \centering
    \begin{minipage}[t]{\textwidth}
    \input{Figures/prompt_question_decomposition}
    \end{minipage}
\end{figure*}

\begin{figure*}[ht]
    \centering
    \begin{minipage}[t]{\textwidth}
    \input{Figures/prompt_subquestion_answering}
    \end{minipage}
\end{figure*}

\begin{figure*}[ht]
    \centering
    \begin{minipage}[t]{\textwidth}
    \input{Figures/prompt_final_answer_synthesis}
    \end{minipage}
\end{figure*}

%% file: Figures/prompt_knowledge_graph_extraction.tex
\begin{tcblisting}{conragpromptbox,title={Knowledge graph extraction prompt}}
# Role
You are an evidence-grounded knowledge graph extraction agent.

# Task
Extract explicit graph facts from a passage for retrieval.

# Instructions
- Use `schema` as the preferred label set; create new labels only when needed.
- Use only facts stated in `passage`; do not infer.
- Preserve passage entity names; merge entities only when the passage clearly treats them as identical.
- Split dense passages into separate facts.
- Put relations in directed triples: subject, relation, object.
- Put literal values in attributes, including dates, roles, locations, genres, counts, awards, occupations, nationalities, and categories.
- Include every entity referenced by triples or attributes in `entities`.
- Keep labels concise and retrieval-friendly.

# Output Format
Return valid JSON only. Do not include markdown, citations, explanations, reasoning, or extra keys.
{
  "attributes": {"<entity_name>": ["<attribute_key>: <attribute_value>"]},
  "triples": [["<subject>", "<relation>", "<object>"]],
  "entities": {"<entity_name>": "<entity_type>"}
}

If no explicit graph facts are present, return:
{"attributes": {}, "triples": [], "entities": {}}

# Input
<schema>
{schema}
</schema>

<passage>
{passage}
</passage>
\end{tcblisting}
\vspace{10pt}
\caption{\textbf{Prompt for knowledge graph extraction.}}
\label{fig:prompt_knowledge_graph_extraction}

%% file: Figures/knowledge_graph_schema.tex
\begin{tcblisting}{conragpromptbox,title={Knowledge graph extraction schema}}
{
  "nodes": [
    "person", "character", "organization", "group", "location", "facility", "event", "work", "award",
    "role", "object", "concept", "species", "natural_feature", "natural_phenomenon", "time_period",
    "language", "religion", "ethnicity"
  ],
  "relations": [
    "located_in", "part_of", "contains", "adjacent_to", "borders", "connected_to", "crosses",
    "flows_through", "born_in", "died_in", "buried_in", "lives_in", "citizen_of", "speaks",
    "follows_religion", "belongs_to_ethnic_group", "holds_role", "works_for", "studies_at",
    "trained_under", "member_of", "affiliated_with", "leads", "controls", "founded", "owns", "acquired",
    "headquartered_in", "capital_of", "parent_of", "sibling_of", "married_to", "friend_of", "created_by",
    "directed_by", "written_by", "composed_by", "produced_by", "published_by", "broadcast_by",
    "performed_by", "acts_in", "appears_in", "portrays", "adapted_from", "based_on", "influenced_by",
    "collaborates_with", "set_in", "depicts", "references", "sequel_to", "prequel_to", "remake_of",
    "wins", "nominated_for", "organizes", "participates_in", "represents", "plays_for", "coaches",
    "competes_against", "defeats", "precedes", "succeeds", "replaces", "allied_with", "supports",
    "opposes", "meets", "attacks", "kills", "protects", "captures", "escapes_from", "betrays",
    "commands", "travels_to", "uses", "seeks", "finds", "reveals", "causes", "named_after",
    "originates_from", "developed_by", "discovered_by", "invented_by", "related_to", "features",
    "designed_by"
  ],
  "attributes": [
    "alias", "former_name", "nickname", "acronym", "description", "definition", "type", "status",
    "occupation", "nationality", "gender", "genre", "language", "religion", "ethnicity", "birth_date",
    "death_date", "cause_of_death", "age", "height", "weight", "position", "rank", "office",
    "term_start", "term_end", "years_active", "debut", "date", "year", "time", "start_date", "end_date",
    "release_date", "publication_date", "runtime", "duration", "season", "edition", "budget", "revenue",
    "box_office", "score", "rating", "count", "population", "area", "size", "length", "elevation",
    "capacity", "value", "prize", "award_count", "founding_date", "dissolution_date", "scientific_name",
    "episode_count", "track_count", "page_count", "coordinates", "color", "trait", "ability", "motive",
    "goal", "appearance", "material", "movement", "style", "instrument", "motto", "website"
  ]
}
\end{tcblisting}
\vspace{10pt}
\caption{\textbf{Schema for knowledge graph extraction.}}
\label{fig:knowledge_graph_schema}

%% file: Figures/prompt_question_decomposition.tex
\begin{tcblisting}{conragpromptbox,title={Question decomposition prompt}}
# Role
You are a retrieval planner for multi-hop question answering.

# Task
Create the retrieval plan required to answer the question using the evidence.

# Instructions
- `acquired_information`: concise facts explicitly stated in `evidence`; no inference.
- Do not answer directly. `plan` must be non-empty.
- Plan only for missing facts or facts that need verification needed to answer `question`.
- Preserve the target slot, entity constraints, comparisons, and yes/no/all/both conditions.
- Use the fewest retrieval-oriented steps; resolve bridge entities before dependent facts.
- Make `sub_question` self-contained except for `<dep:ID>` placeholders.
- Use zero-based contiguous ids; dependencies must be earlier ids.
- Every `<dep:ID>` must appear in `dependencies`; no unused dependencies.

# Output Format
Return valid JSON only. Do not include explanations, markdown, citations, or extra keys.
{
  "acquired_information": "<complete concise grounded facts useful for answering question>",
  "plan": [
    {
      "id": 0,
      "sub_question": "<retrieval-oriented subquestion>",
      "dependencies": [<integer>, ...]
    }
  ]
}

# Input
<question>
{question}
</question>

<evidence>
{evidence}
</evidence>
\end{tcblisting}
\vspace{10pt}
\caption{\textbf{Prompt for question decomposition.}}
\label{fig:prompt_question_decomposition}

%% file: Figures/prompt_subquestion_answering.tex
\begin{tcblisting}{conragpromptbox,title={Subquestion answering prompt}}
# Role
You are an evidence-grounded subquestion answering agent.

# Task
Answer one plan step from evidence.

# Instructions
- Use `original_question` to preserve the target slot and answer type.
- Use `acquired_information` as already grounded context.
- Use `sub_question` as the immediate question to answer.
- Return the shortest grounded value that answers `sub_question`.
- If one entity or value is requested, return one entity or value, not a list.
- Do not substitute related slots such as employer for school, birthplace for nationality, producer for director, or participant for winner.
- Put complete, concise, grounded facts that directly help answer `original_question` in `acquired_information`.

# Output Format
Return valid JSON only. Do not include reasoning traces, markdown, citations, or extra keys.
{
  "answer": "<short answer>",
  "acquired_information": "<complete concise grounded facts useful for answering original_question>"
}

# Input
<original_question>
{original_question}
</original_question>

<acquired_information>
{acquired_information}
</acquired_information>

<sub_question>
{sub_question}
</sub_question>

<evidence>
{evidence}
</evidence>
\end{tcblisting}
\vspace{10pt}
\caption{\textbf{Prompt for subquestion answering.}}
\label{fig:prompt_subquestion_answering}

%% file: Figures/prompt_final_answer_synthesis.tex
\begin{tcblisting}{conragpromptbox,title={Final answer synthesis prompt}}
# Role
You are the final answer agent for an evidence-grounded RAG QA system.

# Task
Produce the direct final answer from the supplied evidence.

# Instructions
- Return only the direct answer, as concisely as possible.
- Do not explain or provide any additional context.
- If the answer is a simple yes/no, return exactly `Yes.` or `No.`
- If the answer is a name, return only the name.
- If the answer is a date, return only the date.
- If the answer is a number, return only the number.
- If the answer requires a brief phrase, make it as concise as possible.
- Give only the essential answer, nothing more.

# Output Format
Return only the final answer string.

# Input
<question>
{question}
</question>

<evidence>
{evidence}
</evidence>
\end{tcblisting}
\vspace{10pt}
\caption{\textbf{Prompt for final answer synthesis.}}
\label{fig:prompt_final_answer_synthesis}